\documentclass[lettersize,journal]{IEEEtran}
\usepackage{amsmath,amsfonts}
\usepackage{algorithmic}
\usepackage{algorithm}
\usepackage{array}
\usepackage[caption=false,font=normalsize,labelfont=sf,textfont=sf]{subfig}
\usepackage{textcomp}
\usepackage{stfloats}
\usepackage{url}
\usepackage{verbatim}
\usepackage{graphicx}
\usepackage{cite}
\usepackage{booktabs}
\usepackage{multirow}
\usepackage{tikz}
\usepackage{booktabs}
\usepackage{makecell}
\usepackage{array}
\usepackage{booktabs}
\usepackage{makecell}
\usepackage{pifont}
\usepackage{tabularx}
\usepackage{pifont}
\usepackage{xcolor}
\providecommand{\cmark}{\ding{51}}

\usetikzlibrary{calc}
\begin{document}

\title{Safety4DM: A Vision-Language Framework for Risk-Aware Dynamic Maps in Vehicle–Road–Cloud Collaborative Driving}

\author{Zhengshu Zhou, Shaoyu Wang, Lu Tao, Xinhui Hu, Jiaqi Gu, and Tingting Zhao
\thanks{Manuscript received xx; revised xx. (Corresponding author: Lu Tao.)}%
\thanks{Zhengshu Zhou, Shaoyu Wang, Xinhui Hu, Jiaqi Gu, and Tingting Zhao are with the College of Artificial Intelligence, Tianjin University of Science and Technology, Tianjin, China (e-mail: zhouzhengshu@tust.edu.cn; 24836940@mail.tust.edu.cn; huxinhui09@mail.tust.edu.cn; gujiaqi@mail.tust.edu.cn; tingting@tust.edu.cn)}
\thanks{Lu Tao is with the School of Resource and Environmental Sciences, Wuhan University, Wuhan, China (ltao@whu.edu.cn).}}

\markboth{Journal of \LaTeX\ Class Files,~Vol.~14, No.~8, August~2021}%
{Shell \MakeLowercase{\textit{et al.}}: A Sample Article Using IEEEtran.cls for IEEE Journals}


\maketitle

\begin{abstract}
Dynamic map (DM) is a core information infrastructure for vehicle-road-cloud (VRC) cooperative autonomous driving. It fuses perception data from multiple sources to overcome the inherent line-of-sight and field-of-view limits of single-vehicle perception. However, DM still falls short on driving safety in two ways. First, DM data specifications already define road conditions semantically, but no pipeline yet turns real-world road observations into fine-grained risk entities in the DM. Second, existing DM systems lack a natural language interface for driving risks. To close these gaps, we propose Safety4DM, a safety-oriented language system for VRC dynamic maps. We first build RoadRisk, a real-world road image dataset that covers four types of road conditions and four risk levels. On this dataset, we fine-tune a vision-language model (VLM) for road safety risk perception. The model outputs structured risk representations that the DM can ingest directly. We then introduce a Safety-aware Step-wise Prompting (SafeStep) mechanism as the natural language query interface of the DM. SafeStep answers queries accurately and also generates actionable driving advice. To build and evaluate this interface, we develop SafeVRC-Sim, a VRC cooperative simulation framework, and SafeDM-QA, a safety question-answering benchmark. Experiments show that even the compact Qwen3-VL-4B reaches 89.83\% accuracy on risk perception after fine-tuning. SafeStep also plugs easily into multiple LLM families. With medium and large models (20B and above) from five families, it achieves over 93\% natural language query accuracy in every case, which shows strong generalization.

\end{abstract}


\begin{figure}
  \centering
  \includegraphics[width=0.48\textwidth]{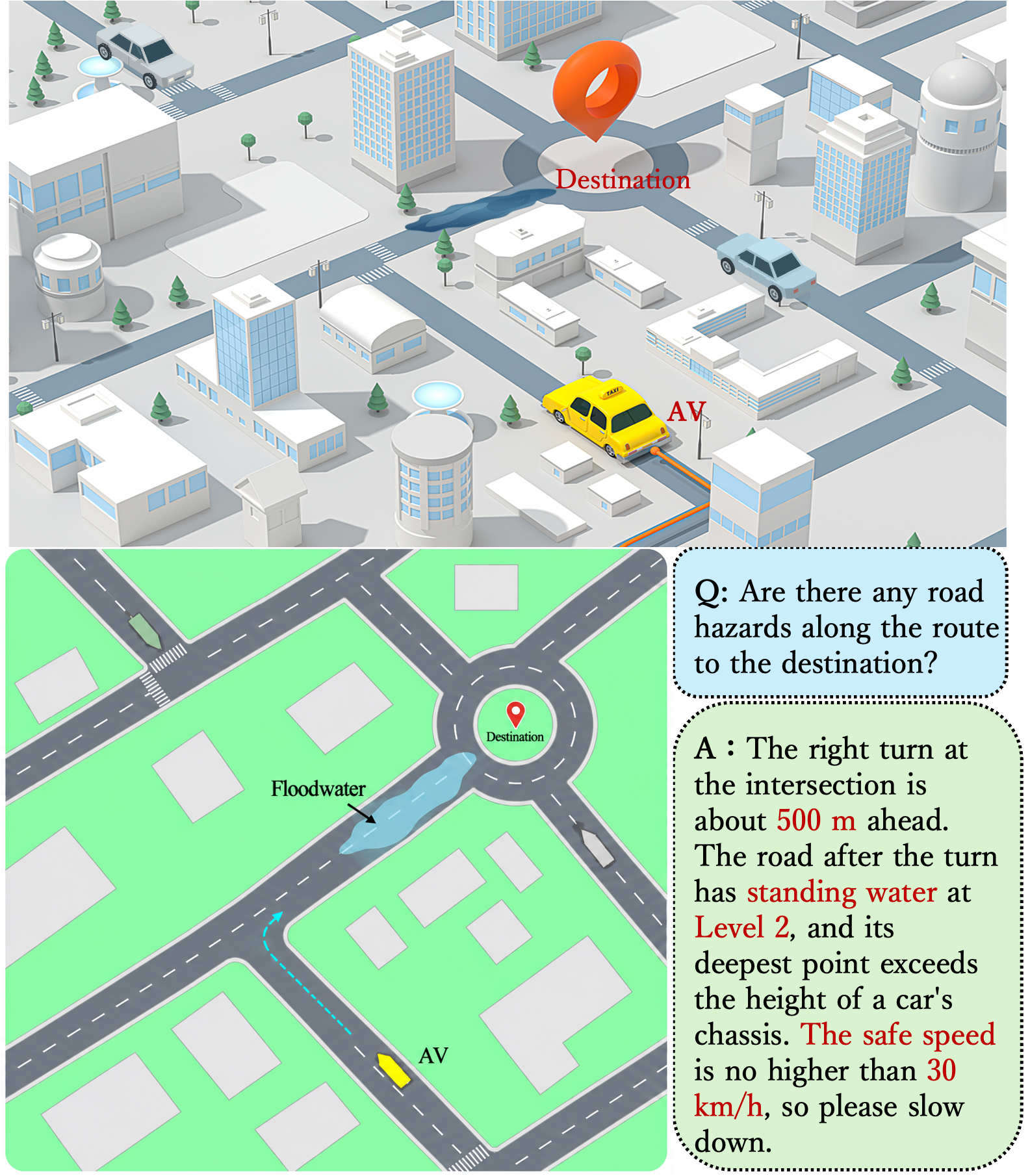}
  \caption{Motivating example. Using a vehicle–road–cloud dynamic map, the AV queries hazards beyond its line of sight and receives the risk and driving advice in natural language.}
  \label{fig_1}
\end{figure}

\section{Introduction}
\IEEEPARstart{I}{n} recent years, large language models (LLMs) have greatly improved how autonomous driving systems understand scenes, reason, and interact with humans, making these systems smarter and more efficient \cite{zhou2024vision}. Safe driving decisions, however, start with accurate and complete perception of the surroundings \cite{luo2024delving}. Unfortunately, most studies on LLM-enhanced autonomous driving (LLM-AD) focus on a single-vehicle view \cite{chiu2026v2v,choudhary2024talk2bev,deruyttere2019talk2car}. Single-vehicle perception has inherent limits that are hard to overcome: its sensing range is short, its view is easily blocked, and it rarely captures the full traffic picture. These limits put driving safety at risk. Vehicle-road-cloud (VRC) cooperation emerged to address this problem. Through cooperative perception across vehicles, roadside units, and the cloud, a VRC system can see much farther and wider than any single vehicle, which makes it very promising for safer autonomous driving \cite{gao2024vehicle}. Yet the dynamic map (DM), the core information infrastructure of VRC cooperation, still has two key gaps in its safety capability.

DM carries cooperative perception data. It gathers the traffic information sensed by vehicles, roads, and the cloud into a structured global map that updates in real time \cite{watanabe2020dynamicmap}. This gives beyond-line-of-sight safety decisions a unified information base. Compared with the ego-centric bird's-eye view (BEV) \cite{ding2024holistic}, the DM offers a God's Eye View (GEV) that covers a much larger area \cite{tao2024enhancing}. DM data specifications already reserve a semantic layer for road condition information \cite{Shimada2015ImplementationAE}. In practice, however, this information comes mainly from dedicated sensors or manual reports. Both options are costly and hard to scale across large urban road networks. With advances in deep learning, image classification and semantic segmentation methods can now detect road hazards such as standing water and snow \cite{levering2020detecting,zhang2024agsenet}. But these methods output only class labels or pixel masks with no explicit reasoning, so they offer no traceable basis for safety-critical risk judgments. Recent work has also brought multimodal large models into road risk recognition \cite{ding2025urban}, but these models cannot yet assess fine-grained risk across multiple road conditions. As a result, there is still no pipeline from real-world road observations to fine-grained, interpretable risk entities in the DM. This is the first gap.

DM builds on a relational model, so SQL-like languages can handle standardized queries and data operations efficiently. This standardization brings efficiency, but it costs flexibility. To this day, the DM exists mainly as structured data and lacks a natural language interface for driving safety. Drivers find it hard to query, understand, and use its risk information directly for beyond-line-of-sight decisions. This is exactly where a natural language interface adds value. Users need no special query syntax, and the interface can draw on the commonsense knowledge of LLMs to turn numerical query results into actionable driving advice. Current DM platforms built on structured queries offer none of these benefits. This is the second gap.

To close these two gaps, we propose Safety4DM, a language system for VRC dynamic maps that targets driving safety. For the first gap, we build RoadRisk, a real-world road image dataset, and use it to train a VLM for road condition risk assessment. The model takes real road images as visual input and outputs structured risk information, such as road condition type and risk level. We then inject this information into the DM. This completes the pipeline from real-world visual risk perception to structured DM updates. For the second gap, we propose a Safety-aware Step-wise Prompting (SafeStep) mechanism as the natural language query interface of the DM. SafeStep answers queries accurately and also generates actionable driving advice. To build and evaluate this interface, we develop SafeVRC-Sim, a VRC cooperative simulation framework. It generates language scenarios that contain both vehicle risks and road condition risks. On top of this framework, we build SafeDM-QA, a safety question-answering benchmark with 1,600 mixed-traffic scenarios and 23.5K question-answer pairs. We use this benchmark to evaluate Safety4DM comprehensively.

Our main contributions are as follows:

\begin{itemize}
    \item We propose \textbf{Safety4DM} to close the gaps of VRC dynamic maps in road risk perception and natural language safety interaction. To the best of our knowledge, it is the first system that brings road environment risks into VRC dynamic maps and supports natural language safety queries on both vehicle risks and road environment risks. It connects the full pipeline from real-world visual perception, to structured risk representation, to natural language safety queries.
    \item We build \textbf{RoadRisk} to train a large model for road safety risk perception. The dataset contains real-world road images that cover four types of road conditions and four risk levels. The model fine-tuned on RoadRisk uses explicit visual reasoning to judge road condition type and risk level at a fine-grained level. Its judgments are traceable, and its outputs are structured risk representations that the DM can ingest directly.
    \item We design \textbf{SafeVRC-Sim} to simulate realistic traffic scenarios under VRC cooperative perception. Based on it, we build SafeDM-QA, a safety question-answering benchmark with 18 safety-related query tasks on vehicle and road environment information.
    \item We propose \textbf{SafeStep}, a step-wise prompting query mechanism for the DM. SafeStep offloads the computation of spatial relations and safety metrics to a query toolbox. This lets LLMs answer natural language safety queries precisely over structured scenes and draw on commonsense knowledge to generate actionable driving advice.
\end{itemize}

% ===== Preamble (if not already defined) =====
% \newcommand{\cmark}{\ding{51}}
% \newcommand{\pmark}{$\triangle$}

% ===== Table code =====
\begin{table*}[t]
\renewcommand\arraystretch{1.1}
\centering
\caption{Comparison with related studies in LLM-enhanced autonomous driving.}
\label{tab:related_studies}
\scriptsize
\setlength{\tabcolsep}{4pt}
\resizebox{\textwidth}{!}{
\begin{tabular}{l c c c c c c c c c}
\toprule
\multirow{2}{*}{\textbf{Model}} &
\multirow{2}{*}{\textbf{Publication}} &
\multicolumn{2}{c}{\textbf{Setting}} &
\multicolumn{4}{c}{\textbf{Task coverage}} &
\multirow{2}{*}{\makecell{\textbf{Scene}\\\textbf{elements}}} &
\multirow{2}{*}{\textbf{Output}} \\
\cmidrule(lr){3-4}
\cmidrule(lr){5-8}
 & &
\makecell{agent\\setting} &
\makecell{data\\source} &
\makecell{perce-\\ption} &
\makecell{predic-\\tion} &
\makecell{plan-\\ning} &
\makecell{NL\\query} &
 & \\
\midrule
DriveGPT4~\cite{xu2024drivegpt4} &
IEEE RAL 2024 & Single & Real &
\cmark &  & \cmark & \cmark &
OI & Text + Control \\
DriveVLM~\cite{tian2024drivevlm} &
CoRL 2024 & Single & Real &
\cmark & \cmark & \cmark &  &
OI + EI & Text + Trajectory \\
DriveLM~\cite{Sima2023DriveLMDW} &
ECCV 2024 & Single & Sim + Real &
\cmark & \cmark & \cmark & \cmark &
OI & Text + Trajectory \\
Talk2BEV~\cite{choudhary2024talk2bev} &
ICRA 2024 & Single & Real &
\cmark &  &  & \cmark &
OI & Text \\
PromptTrack~\cite{Wu2023LanguagePF} &
AAAI 2025 & Single & Real &
\cmark &  &  &  &
OI & 3D boxes \\
NuScenes-QA~\cite{qian2024nuscenes} &
AAAI 2024 & Single & Real &
\cmark &  &  & \cmark &
OI & Text \\
LMDrive~\cite{shao2024lmdrive} &
CVPR 2024 & Single & Sim &
 &  & \cmark &  &
OI & Control \\
V2X-VLM~\cite{you2026v2x} &
TR-C 2026 & V2X & Real &
 &  & \cmark &  &
OI & Trajectory \\
V2V-LLM~\cite{chiu2026v2v} &
ICRA 2026 & V2V & Real &
\cmark &  & \cmark & \cmark &
OI & Text + Trajectory \\
\midrule
\textbf{Safety4DM [this]} &
-- & VRC & Sim + Real &
\cmark &  &  & \cmark &
OI + EI & Text + Risk entities \\
\bottomrule
\end{tabular}
}

\par\vspace{5pt}
\noindent
\begin{minipage}{\textwidth}
\footnotesize
Single: single vehicle; V2V / V2X: vehicle-to-vehicle / vehicle-to-everything;
VRC: vehicle-road-cloud.
OI: vehicle information; EI: environment information.
\end{minipage}

\end{table*}

\begin{figure}[h]
  \centering
  \includegraphics[width=0.48\textwidth]{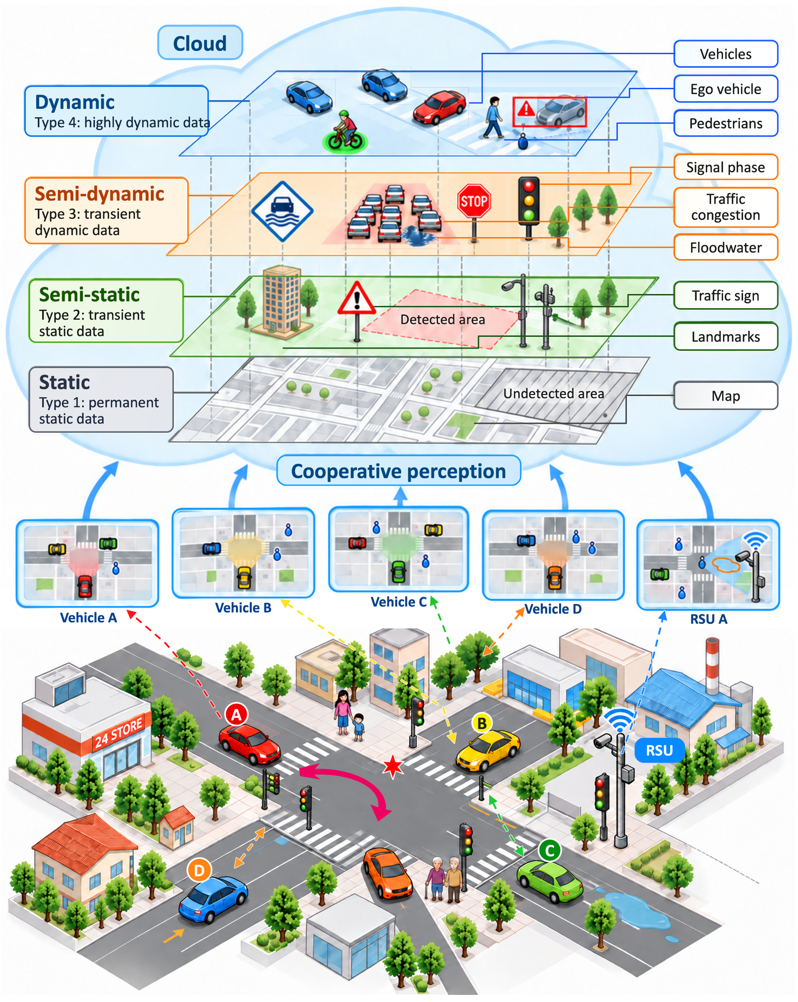}
  \caption{Four-layer structure of the dynamic map. Vehicles and RSUs upload their perception results to the cloud through cooperative perception, where the information is organized into static, semi-static, semi-dynamic, and dynamic layers by update frequency.}
  \label{fig_DM}
\end{figure}

\section{Related Work}
\subsection{LLM-Enhanced Autonomous Driving}
LLMs have strong reasoning abilities and rich pretrained knowledge. Researchers have therefore applied them to a wide range of autonomous driving tasks, including scene understanding and perception enhancement, driving question answering and reasoning, prediction, decision-making and planning, end-to-end driving and control, and natural language interaction and cooperative queries. Early work in this area mainly tested whether large models could support driving perception, decision-making, and explanation. For perception and understanding, Talk2BEV adds a language interface to bird's-eye view representations \cite{choudhary2024talk2bev}, so users can query and reason about traffic scenes in natural language. BEV-InMLLM fuses instruction-aware BEV features with a multimodal large model \cite{ding2024holistic}, allowing it to understand traffic scenes according to natural language instructions. For decision-making and explanation, GPT-Driver was the first to recast motion planning as a language reasoning problem \cite{mao2023gpt}. DriveGPT4 explains its driving decisions in natural language and directly outputs low-level control signals, achieving interpretable end-to-end driving \cite{xu2024drivegpt4}. Meanwhile, benchmarks such as NuScenes-QA \cite{qian2024nuscenes}, LingoQA \cite{marcu2024lingoqa}, and Reason2Drive \cite{nie2024reason2drive} appeared one after another. They moved evaluation beyond basic question answering toward high-level scene understanding and interpretable reasoning. Research focus also shifted toward closed-loop driving and long-tail understanding: LMDrive achieves closed-loop end-to-end control \cite{shao2024lmdrive}, and DriveVLM strengthens long-tail scene understanding and hierarchical planning \cite{tian2024drivevlm}. Together, these studies have greatly improved scene understanding, interpretability, and interaction flexibility in autonomous driving.

More recently, research has expanded from the single-vehicle view to a cooperative view. UniV2X built the first end-to-end vehicle-infrastructure cooperative driving framework, unifying cross-view perception, mapping, and planning in one network. V2X-VLM goes further and introduces a vision-language model \cite{you2026v2x}. It fuses vehicle-side and roadside images with textual scene descriptions to perform end-to-end cooperative trajectory planning. V2V-LLM uses a centralized large model to fuse perception data from multiple vehicles \cite{chiu2026v2v}. Each vehicle can then ask driving safety questions in natural language, and the authors also built the V2V-QA benchmark for this task. Talk2DM, a precursor to this work, targets VRC dynamic maps and takes a first step toward LLM-based spatial queries about surrounding vehicles \cite{Tao2026Talk2DMEN}. Unlike these studies, we build a driving safety framework for dynamic maps. It brings the perception of real-world road environment risks and safety question answering that couples vehicles with their environment into one unified pipeline, a direction the literature has not yet explored. Table \ref{tab:related_studies} compares our work with related studies.

\subsection{Dynamic Maps}

Dynamic map (DM) is a shared database indexed by time and space. It overlays constantly changing traffic information on a static high-definition map, so traffic participants can access road information beyond their own sensing range. The concept comes from the local dynamic map (LDM) proposed by the European SAFESPOT project. ETSI and ISO later standardized it, and Japan's SIP-adus program adopted it as the core information infrastructure for cooperative autonomous driving \cite{etsi2014ldm}. As Fig. \ref{fig_DM} shows, DM sorts information into four layers by update frequency. The static layer updates about once a month and holds data such as lane-level HD maps. The semi-static layer updates within an hour and covers traffic control plans and road works. The semi-dynamic layer updates within a minute and covers accidents, congestion, local weather, and road conditions. The dynamic layer updates within a second and covers vehicles, pedestrians, and signal phases. The semi-dynamic layer already reserves a slot for road condition information, and ETSI's DENM messages also define event types such as low road adhesion from snow or ice and obstacles on the road. In this paper, vehicle information belongs to the dynamic layer, and the road risk entities from our perception model fill the semi-dynamic layer.

In terms of system implementation, DynamicMap 2.0 deploys the DM across three tiers: vehicle, edge, and cloud. The edge tier shares local real-time information, for example to assist vehicles at intersections  \cite{Hosono2021ImplementationAE}. The cloud tier handles wide-area, non-real-time tasks such as citywide traffic optimization. Fusing data from multiple sensors lets them cover each other's blind spots. For data management, the DM stores static maps in relational form and processes dynamic information in memory as timestamped data streams. Applications access the DM through an SQL-like continuous query language. This kind of structured interface is efficient, but it targets developers: users must know the data schema, spatial predicates, and query syntax.

\subsection{Road Environment Risk Perception}

Road environment risk perception aims to identify hazards in the road environment that threaten driving safety, such as standing water, snow, ice, potholes, and obstacles, and to assess how dangerous they are. It lays the foundation for safe driving decisions. Unlike risks from dynamic traffic participants, these risks come from the road environment itself. Many studies have addressed this task. Traditional methods rely on sensors and other hardware, which are expensive and cannot cover city streets at scale \cite{schenato2021rugged,moy2017floodx}. With the rise of deep learning, more and more researchers have turned to deep learning methods for this task. Levering et al. were among the first to study road risks systematically\cite{levering2020detecting}. They used a ResNet-34 image classifier to recognize eight types of events on the road: animals on the road, road collapse, vehicle accidents, fire, flooding, landslides, snow, and fallen trees. Later research shifted toward fine-grained perception of specific road surface risks. AGSENet uses semantic segmentation to detect standing water on roads under complex weather and lighting conditions \cite{zhang2024agsenet}. Datasets such as ACDC\cite{sakaridis2025acdc} and IDD-AW\cite{Shaik2023IDDAWAB} further study road scene understanding and safety robustness in rain, fog, snow, and at night.

\section{Method}

\begin{figure*}
  \centering
  \includegraphics[width=\textwidth]{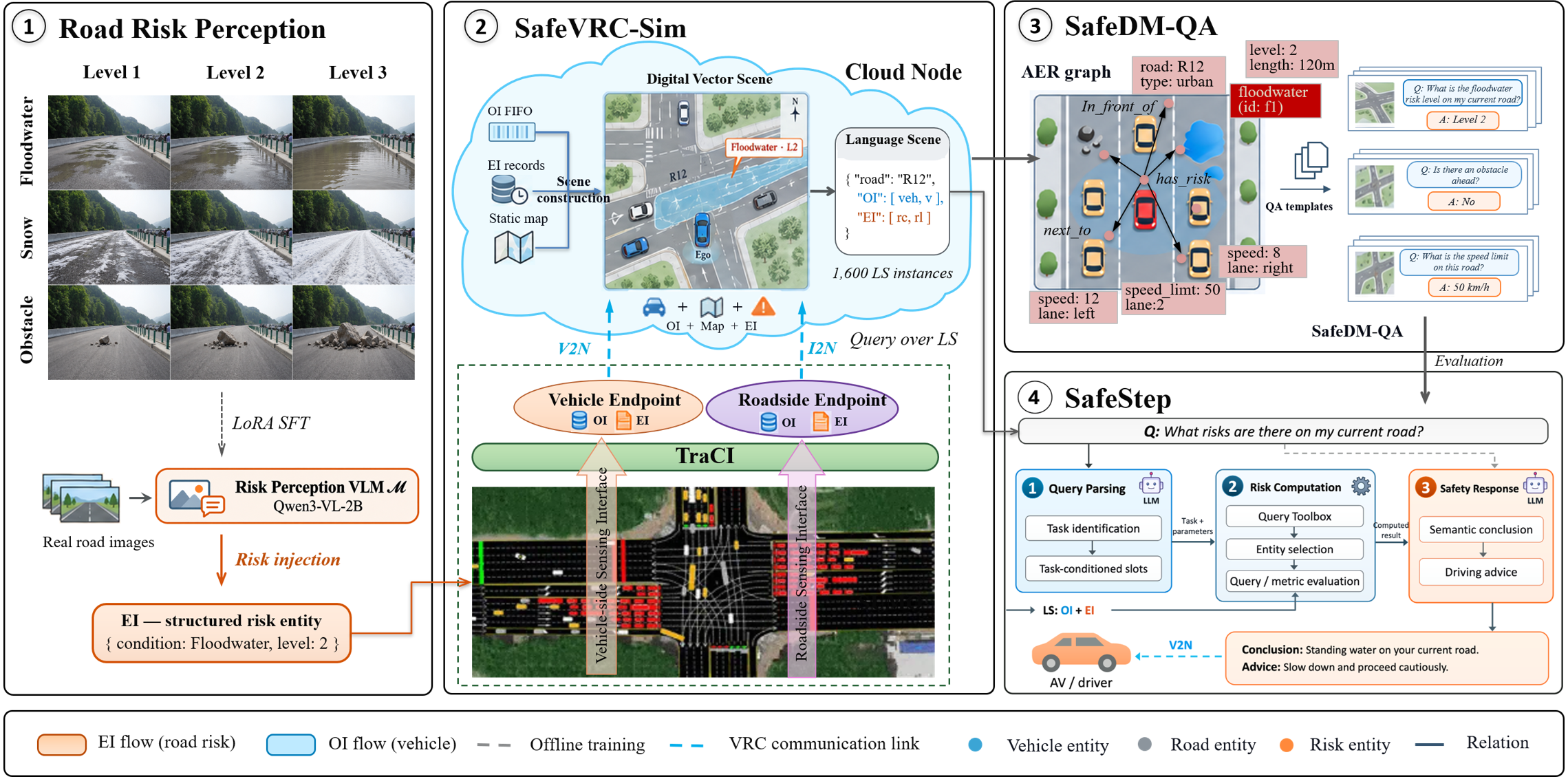}
  \caption{Overall architecture of Safety4DM. The fine-tuned VLM turns road images into risk entities, SafeVRC-Sim fuses them with vehicle states into a language scene, and SafeStep answers safety queries over it.}
  \label{fig_2}
\end{figure*}

Safety4DM forms a closed loop from road risk perception on real images to natural language safety queries. This section presents its five key components: the road safety risk perception model, the road safety risk dataset that supports it, the SafeVRC-Sim framework for simulating VRC cooperation, the SafeDM-QA dataset for safety question answering in both the OI and EI domains, and the SafeStep mechanism that guides LLMs through natural language safety queries. Fig.\ref{fig_2} shows the overall architecture.

% ===== Table =====
\begin{table}[t]
\centering
\caption{Definition of road-condition categories and risk levels.}
\label{tab:risk_levels}
\footnotesize
\setlength{\tabcolsep}{5pt}
\renewcommand{\arraystretch}{1.3}
\resizebox{0.4\textwidth}{!}{
\begin{tabular}{l c p{0.62\columnwidth}}
\hline\hline
\textbf{Condition} & \textbf{Level} & \textbf{Decision criterion} \\
\hline
Normal & 0 &
no  floodwater, no snow cover, no obstacle \\
\hline
\multirow{3}{*}{Floodwater} & 1 &
Mild: deepest point $<$15\,cm \\
\cmidrule(l){2-3}
 & 2 &
Moderate: deepest point 15--30\,cm; \\
\cmidrule(l){2-3}
 & 3 &
Severe: deepest point $>$30\,cm; \\
\hline
\multirow{3}{*}{Snow} & 1 &
Drivable lane $<$50\% snow-covered; \\
\cmidrule(l){2-3}
 & 2 &
Lane 50--100\% covered; mean depth $<$10\,cm. \\
\cmidrule(l){2-3}
 & 3 &
Lane 80--100\% covered; mean depth $>$10\,cm. \\
\hline
\multirow{3}{*}{Obstacle} & 1 &
Lateral span $<$1/4 lane width \\
\cmidrule(l){2-3}
 & 2 &
Lateral span 1/4--2/3 lane width \\
\cmidrule(l){2-3}
 & 3 &
Lateral span $>$2/3 lane or fully across \\
\hline\hline
\end{tabular}
}
\vspace{2pt}
\end{table}

\begin{table}
\centering
\caption{Diversity Analysis of the RoadRisk Dataset.}
\label{tab:roadrisk}
\begin{tabular}{clr}
\hline\hline
\textbf{Dimension} & \textbf{Category} & \textbf{Number} \\
\hline
\multirow{6}{*}{Scene}
 & Urban streets & 2423 \\
 & Highways      & 1373 \\
 & Alleys        & 732 \\
 & Villages      & 505 \\
 & Tunnels       & 502 \\
 & Towns         & 465 \\
\cmidrule(l){2-3}
\multirow{2}{*}{Illumination}
 & Daytime   & 4364 \\
 & Nighttime & 1636 \\
\cmidrule(l){2-3}
\multirow{4}{*}{Viewpoint}
 & Dashcam    & 2326 \\
 & CCTV       & 2103 \\
 & Pedestrian & 1078 \\
 & UAV        & 493 \\
\hline\hline
\end{tabular}
\end{table}

\subsection{Road Safety Risk Perception}
\label{Risk_Perception}

\textbf{Task Definition:}
We formulate road safety risk perception as a conditional generation problem with explicit visual reasoning. Let $\mathcal{D} = \{(I_i, Q_i, y_i)\}_{i=1}^{N}$ denote the road safety risk dataset, where $I \in \mathbb{R}^{H \times W \times 3}$ is the input road image, $Q_i$ is the task-specific instruction, and $y_i$ is the ground-truth risk annotation. Each annotation is a structured risk description $y = (c, r)$, where $c \in \mathcal{C}$ is the road condition category with $\mathcal{C} = \{\text{Normal}, \text{Floodwater}, \text{Snow}, \text{Obstacle}\}$, and $r \in \{0, 1, 2, 3\}$ is the matching risk level. A vision encoder and a text embedding layer map the road image and the task instruction respectively, and we concatenate the two to form a unified context representation for generation:
\begin{equation}
    C = \left[\mathcal{E}_{\mathrm{vision}}(I);\, \mathcal{E}_{\mathrm{text}}(Q)\right]
    \label{eq:context}
\end{equation}
We model the joint probability of the reasoning chain $K = \{k_1, \dots, k_m\}$ and the final answer $A$, and generate the structured sequence $S = (K, A)$:
\begin{equation}
    P(S \mid I, Q) = P(A \mid C, K) \cdot \prod_{t=1}^{|K|} P\left(k_t \mid C, k_{<t}\right)
    \label{eq:generation}
\end{equation}
This formulation forces the model to reason about risk explicitly before it gives the final prediction $A = (\hat{c}, \hat{r})$. The reasoning lives in $K$ and draws on the road surface state and visual references. The model first identifies the basic road condition $\hat{c}$. It then locates key references such as pedestrians, vehicle tires, curbs, lane markings, and wheel ruts. From these references it estimates water depth, snow coverage, or the share of the lane that an obstacle blocks laterally, and finally infers the risk level $\hat{r}$. The reasoning chain has to accommodate the assessment logic and reference system of each road condition within one framework. It also organizes the road condition category and the risk level into a structured output $A$. This output goes directly into the dynamic map as a road environment risk entity and gives the later natural language safety queries a visually grounded source of risk information.

\textbf{Grading Criteria:}
We design the risk levels around how much a road condition actually affects vehicle passage. We treat the drivable road surface as the assessment region and choose, for each condition, a physical quantity that marks the turning point of passage risk. For floodwater we use the water depth at its deepest point. For snow we use lane coverage together with snow thickness. For obstacles we use the share of the lane that the obstacle blocks laterally. Table~\ref{tab:risk_levels} gives the full grading definitions for each road condition.

\begin{figure*}[!t]
  \centering
  \includegraphics[width=0.9\textwidth]{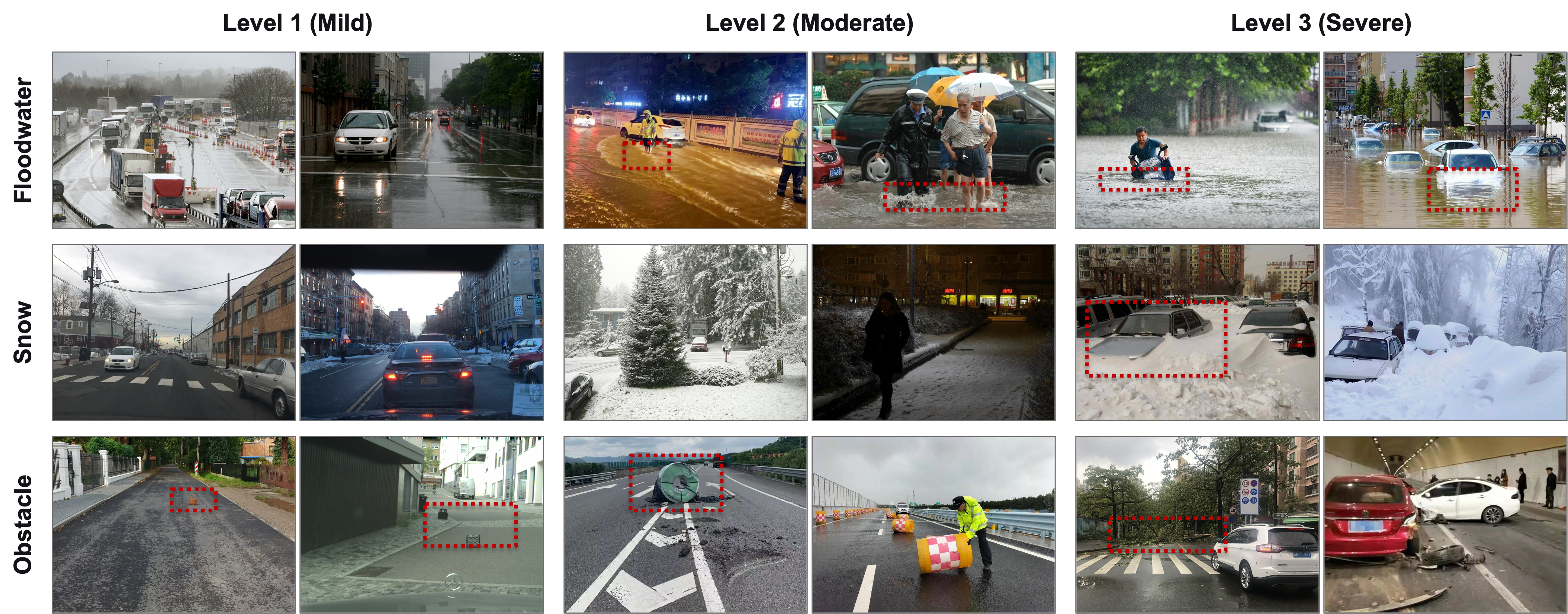}
  \caption{Representative RoadRisk samples for each road condition and risk level. Red dashed boxes mark the key regions used for grading.}
  \label{fig:dataset_samples}
\end{figure*}

% \begin{figure}
%   \centering
%   \includegraphics[width=0.35\textwidth]{figures/fig3.pdf}
%   \caption{The architecture of SafeVRC-Sim}
%   \label{fig:dataset_stats}
% \end{figure}

\subsection{RoadRisk Dataset}

To support SFT training and evaluation of the road safety risk perception LLM, we build our own road risk dataset, RoadRisk. Below we describe how we collect the data, how we annotate it, and how diverse the result is.

\textbf{Data Collection:}
The images in RoadRisk come from two sources. First, we select suitable samples from public datasets. We take normal, floodwater, and snow road images from BDD100K, and obstacle images from Lost and Found, a dataset built specifically for road obstacle detection. Second, we crawl the web for additional floodwater, snow, and obstacle images, which adds diversity from real-world scenes.

\textbf{Data Annotation:}
Annotation runs in two stages: labeling the road condition and risk level, then annotating the reasoning text. In the first stage, professional annotators label the road condition category and risk level of each image according to the risk definitions in Table~\ref{tab:risk_levels}. We apply a strict cross-validation protocol to keep the labels consistent. This stage makes the conclusion labels, which serve as the supervision signal, reliable. In the second stage, we prompt the closed-source Gemini-3-Pro model with a purpose-built system prompt to generate reasoning text for each image. Professional annotators then review and correct this VQA dataset by hand. They check that each reasoning chain follows the grading criteria and agrees with the human conclusion. This process yields RoadRisk, a question-answering dataset of 6{,}000 images. RoadRisk keeps a strict class balance across road conditions and risk levels, which matters for avoiding model bias and for fair evaluation.

\textbf{Diversity Analysis:}
As Table~\ref{tab:roadrisk} shows, RoadRisk covers urban roads, highways, tunnels, and rural and small-town environments. It includes both daytime and nighttime lighting. The images also come from several viewpoints: onboard dashcams, roadside surveillance cameras, pedestrian views, and drones.

\subsection{SafeVRC-Sim}

SafeVRC-Sim serves three purposes. First, it simulates collaborative perception scenarios in the VRC framework and generates structured perception data streams. Second, it injects road environment risks into traffic scenarios. Third, it aggregates multi-source perception data in the cloud, rebuilds the digital traffic scene from this data, and renders the scene as a text representation that feeds LLM question answering. SafeVRC-Sim builds on three open-source platforms: SUMO for microscopic traffic simulation \cite{lopez2018sumo}, ROS2 as a robot operating system for distributed systems\cite{macenski2022ros2}, and PyQt6 as a cross-platform GUI framework \cite{pyqt6}. Fig.~\ref{fig_2} shows its overall architecture.

In SafeVRC-Sim, SUMO simulates realistic traffic flow and generates the traffic data stream. We access this stream through the Traffic Control Interface (TraCI) and build data interfaces for vehicles and RSUs on top of it. With these interfaces, we construct two types of perception nodes: onboard (AV) nodes and roadside (RSU) nodes. Both types perceive vehicle risks and environment risks. For an AV node, the perception result is:

\begin{equation}
    \mathcal{S}_{av} = \big\langle P_{av},\, A_{av} \big\rangle, \quad
    A_{av} = \mathcal{M}\big(I_{av},\, Q\big),
    \label{eq:av_perception}
\end{equation}
\begin{equation}
    P_{av} = \big\{obj \mid \|(x_o,y_o)-(x_e,y_e)\|_2 \leq R_s \big\}.
    \label{eq:av_objects}
\end{equation}

For an RSU node, the perception result is:
\begin{equation}
    \mathcal{S}_{rsu} = \big\langle P_{rsu},\, A_{rsu} \big\rangle, \quad
    A_{rsu} = \mathcal{M}\big(I_{rsu},\, Q\big),
    \label{eq:rsu_perception}
\end{equation}
\begin{equation}
    P_{rsu} = \big\{\mathit{obj} \mid (x_o,y_o) \in L_s,\ L_s = \cup(L_i) \big\}.
    \label{eq:rsu_objects}
\end{equation}

Here $P$ denotes the set of vehicle objects that the node observes, and $obj$ is a perceived object. The pairs $(x_o, y_o)$ and $(x_e, y_e)$ give the coordinates of the detected vehicle and the ego vehicle. The radius $R_s$ is the detection range of the AV sensor, and $L_s$ denotes the lane segments that roadside sensors cover. Road environment risks enter the scene through perception on real images. At a preset frequency, each perception node samples an image $I$ from the real road image library to stand in for the image of the road it occupies. The road safety risk perception LLM $\mathcal{M}$ from Section~\ref{Risk_Perception} then assesses this image, and the resulting risk representation $A = (\hat{c}, \hat{r})$ goes into the scene as the environment information (EI) of that road. We adopt this design because a microscopic traffic simulator carries no visual information about road surface appearance and cannot express road environment risks on its own. Real road images therefore have to come from outside as visual observations, which grounds the risk perception step in real visual data.

Each perception node publishes its data through ROS2, and a cloud node aggregates it. OI and EI differ in how long they stay valid, so the cloud node keeps a separate data structure for each. Vehicle states update often and last only a moment, so we hold the most recent observations in a fixed-length FIFO queue. Road risks update rarely and persist, so we organize them as key-value records indexed by road and attach an expiry time to keep the information current. On top of this, we design a scene constructor that pulls OI and EI from the queue and the key-value records at a preset frequency. It then builds a comprehensive digital vector scene (DVS) over a pre-built static HD map. For efficiency, we decouple the scene constructor from the scene representation, and the latter renders into a visual or a language representation on demand. This work focuses on the language scene (LS). We organize the LS in JSON and pass it to the downstream LLM node for scene understanding and reasoning. It carries the full vehicle states along with the risk category, risk level, and affected lanes of each road. The system also provides a PyQt6 visualization interface, shown in Fig.~\ref{fig_1}. Together, these parts give SafeVRC-Sim a complete pipeline that runs from traffic flow simulation through multi-source collaborative perception and road risk injection to cloud-side scene reconstruction. The pipeline produces 1{,}600 LS instances, which we use to build SafeDM-QA.

\begin{figure}
  \centering
  \includegraphics[width=0.42\textwidth]{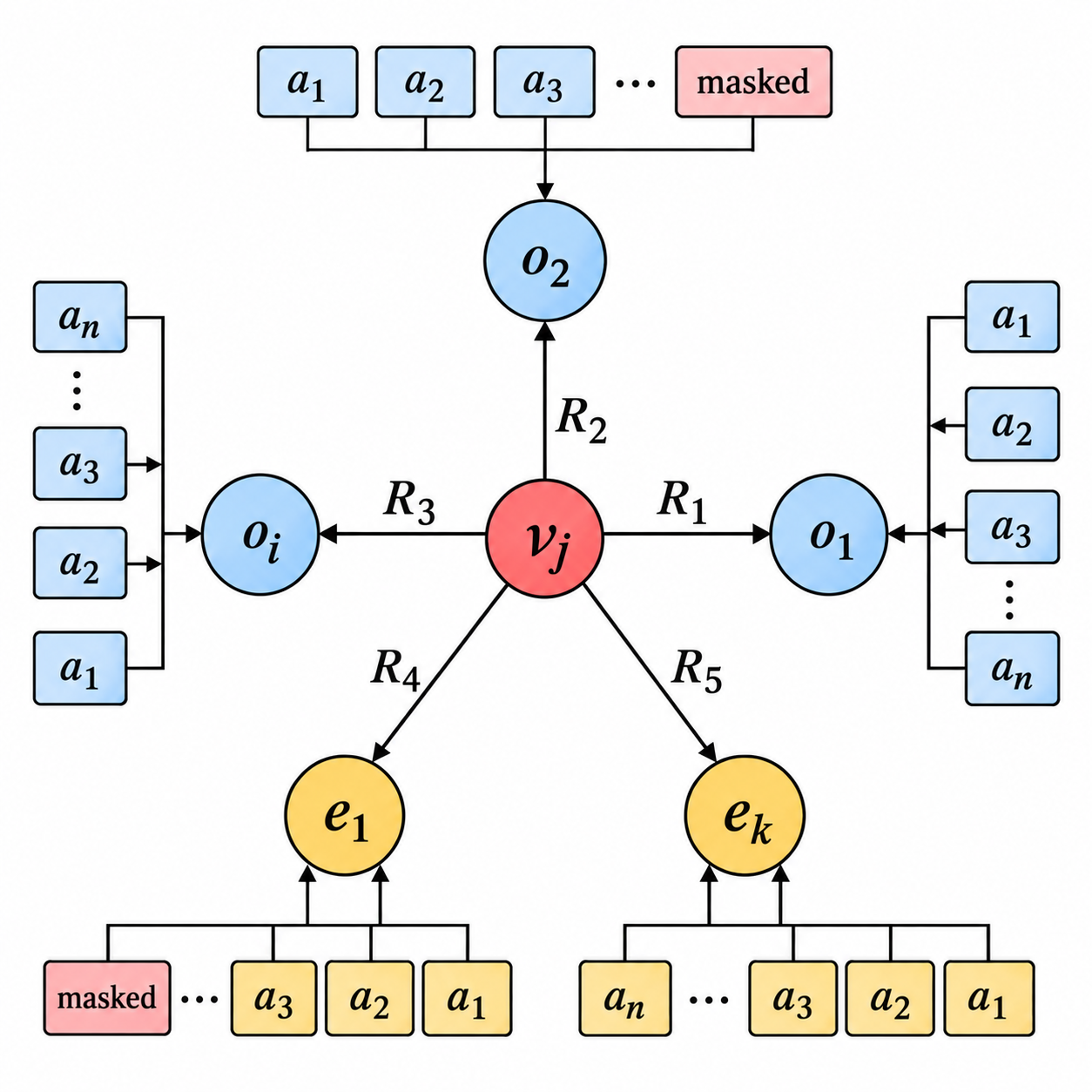}
  \caption{Attribute-entity-relation (AER) graph centered on ego vehicle $v_j$. Surrounding vehicles $o_i$ and road risks $e_k$ connect to $v_j$ via relations $R$; one attribute is masked as the query target.}
  \label{fig_AER}
\end{figure}

\begin{figure}
  \centering
  \includegraphics[width=0.49\textwidth]{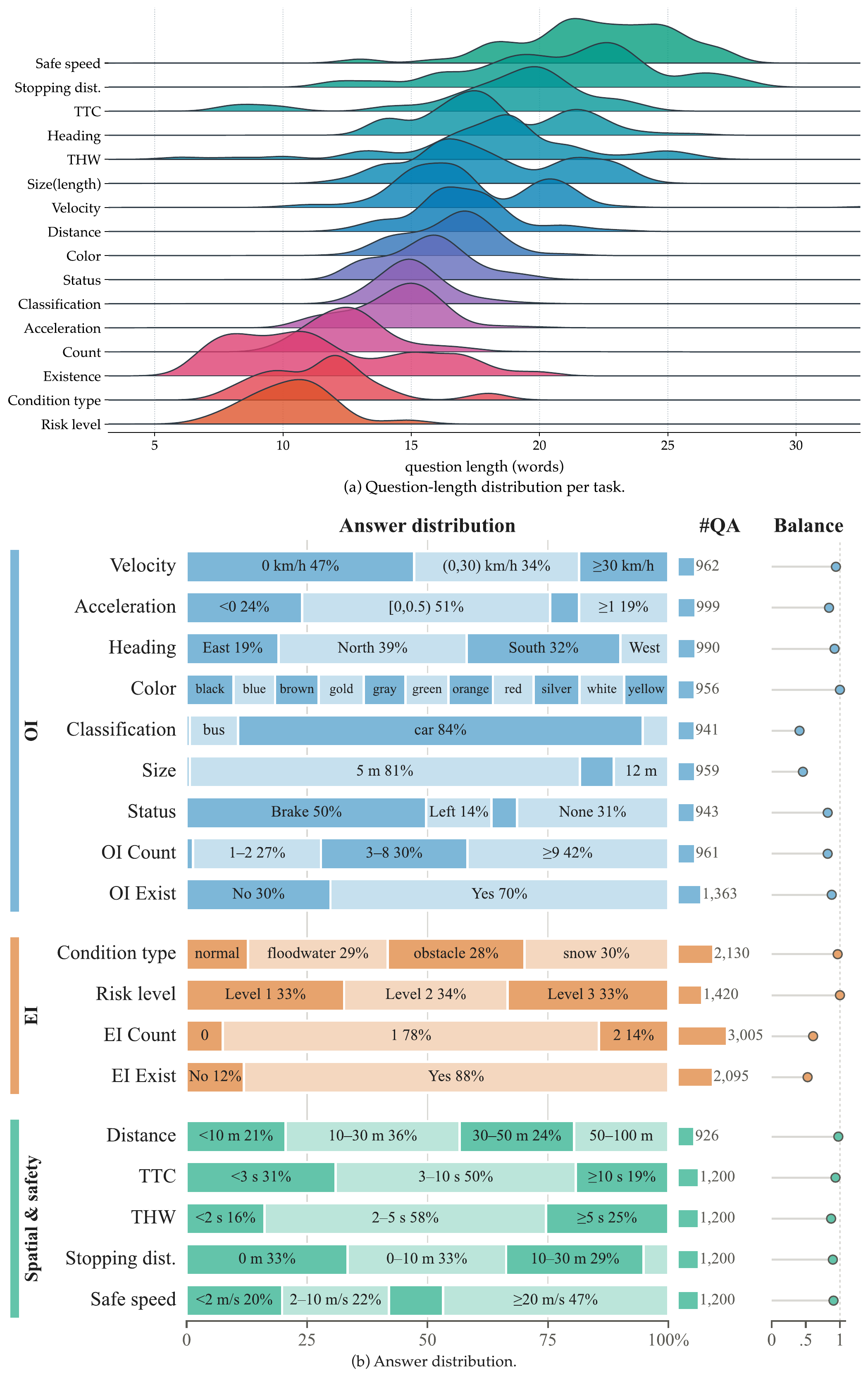}
  \caption{Statistics of SafeDM-QA. (a) Question length per task. (b) Answer distribution and number of QA pairs per task. Balance denotes the normalized Shannon entropy of the answer distribution (1 = perfectly uniform).}
  \label{fig:dataset_stats}
\end{figure}

\begin{table}
\centering
\caption{Schema of object information (OI) and environment information (EI).}
\label{tab:OI_EI_schema}
\begin{tabular}{ll}
\toprule\toprule
\multicolumn{2}{c}{\textbf{OI}} \\
\cmidrule(lr){1-2}
\textbf{Field} & \textbf{Description} \\
\midrule
id & identifier \\
t  & time stamp \\
x  & x coordinate \\
y  & y coordinate \\
v  & velocity \\
a  & acceleration \\
h  & heading \\
le & length of the object \\
wi & width of the object \\
he & height of the object \\
ty & object type \\
co & object color \\
ln & the current lane \\
lx & lane index \\
rd & the current road \\
\bottomrule\bottomrule
\end{tabular}%
\hspace{1em}%      
\begin{tabular}{ll}
\toprule\toprule
\multicolumn{2}{c}{\textbf{EI}} \\
\cmidrule(lr){1-2}
\textbf{Field} & \textbf{Description} \\
\midrule
id & identifier \\
rd & the current road \\
rc & risk category \\
rl & risk level (0--3) \\
af & affected lane \\
ds & the detecting sensor \\
t  & time stamp \\
\bottomrule\bottomrule
\end{tabular}
\end{table}

\begin{table}
\renewcommand\arraystretch{1.3}
\centering
\caption{Query Tasks Defined in SafeStep.}
\label{tab:tasks}
\resizebox{0.5\textwidth}{!}{
\begin{tabular}{llll}
\hline\hline
\textbf{Task} & \textbf{Description} & \textbf{Task} & \textbf{Description} \\
\hline
(1)~~ConditionType  & road risk category      & (10) Size             & vehicle size \\
(2)~~RiskLevel      & road risk level         & (11) Status           & signal status \\
(3)~~EI Count       & number of road risks    & (12) Distance         & distance to ego \\
(4)~~EI Exist       & presence of road risk   & (13) OI Count         & number of vehicles \\
(5)~~Velocity       & vehicle velocity        & (14) OI Exist         & presence of vehicles \\
(6)~~Acceleration   & vehicle acceleration    & (15) TTC              & time to collision \\
(7)~~Heading        & vehicle heading         & (16) THW              & time headway \\
(8)~~Color          & vehicle color           & (17) SafeSpeed        & safe speed under road risk \\
(9)~~Classification & vehicle type            & (18) StoppingDistance & required stopping distance \\
\hline\hline
\end{tabular}
}
\end{table}

% \begin{table}
% \centering
% \caption{Query Tasks Defined in the CoP Task Classification.}
% \label{tab:tasks}
% \resizebox{0.5\textwidth}{!}{
% \begin{tabular}{llll}
% \hline\hline
% \textbf{Task} & \textbf{Description} & \textbf{Task} & \textbf{Description} \\
% \hline
% (1)~~ConditionType  & road risk category      & (10) Size            & vehicle size \\
% (2)~~RiskLevel      & road risk level         & (11) Status          & signal status \\
% (3)~~EI Count       & number of risky roads   & (12) Distance        & distance to ego \\
% (4)~~EI Exist       & presence of road risk   & (13) OI Count        & number of vehicles \\
% (5)~~Velocity       & vehicle velocity        & (14) OI Exist        & presence of vehicles \\
% (6)~~Acceleration   & vehicle acceleration    & (15) TTC             & time to collision \\
% (7)~~Heading        & vehicle heading         & (16) THW             & time headway \\
% (8)~~Color          & vehicle color           & (17) SafeSpeed       & safe speed under road risk \\
% (9)~~Classification & vehicle type            & (18) StoppingDistance & required stopping distance \\
% \hline\hline
% \end{tabular}
% }
% \end{table}

% Packages: \usepackage{booktabs, multirow, makecell}
\begin{table}[t]
\centering
\caption{Question Template Distribution.}
\label{tab:Template}
\footnotesize
\setlength{\tabcolsep}{2pt}
\renewcommand{\arraystretch}{1.1}
\begin{tabular}{@{}l l c r r r@{}}
\toprule
\textbf{Query type} & \textbf{Task} & \textbf{Domain} &
\makecell{\textbf{Ego-}\\\textbf{anchored}} &
\makecell{\textbf{Road-}\\\textbf{anchored}} &
\textbf{Total} \\
\midrule
\multirow{9}{*}{attribute query}
  & velocity         & OI    & 98  & 14 & 112 \\
  & acceleration     & OI    & 143 & 16 & 159 \\
  & heading          & OI    & 132 & 18 & 150 \\
  & color            & OI    & 94  & 12 & 106 \\
  & classification   & OI    & 72  & 19 & 91  \\
  & size             & OI    & 79  & 9  & 88  \\
  & status           & OI    & 80  & 13 & 93  \\
  & condition type   & EI    & 142 & 13 & 155 \\
  & risk level       & EI    & 71  & 14 & 85  \\
\midrule
\multirow{4}{*}{safety query}
  & TTC               & OI+EI & 83 & 13 & 96  \\
  & THW               & OI+EI & 91 & 17 & 108 \\
  & stopping distance & OI+EI & 93 & 18 & 111 \\
  & safe speed        & OI+EI & 91 & 19 & 110 \\
\midrule
\multicolumn{2}{@{}l}{distance query}  & OI    & 86  & 12 & 98  \\
\multicolumn{2}{@{}l}{count query}     & OI/EI & 132 & 13 & 145 \\
\multicolumn{2}{@{}l}{existence query} & OI/EI & 112 & 16 & 128 \\
\midrule
\multicolumn{3}{@{}l}{\textbf{Total}} & \textbf{1599} & \textbf{236} & \textbf{1835} \\
\bottomrule
\end{tabular}
\end{table}

\subsection{SafeDM-QA Data Engineering}
\label{SafeDM-QA}

SafeDM-QA is one of the main contributions of this paper. To the best of our knowledge, it is the first VRC-CP dataset that covers both road environment risk information and dynamic vehicle information, and we tailor it specifically for safety question answering in VRC collaborative driving scenarios. We generate SafeDM-QA through the following pipeline.

\subsubsection{Object Association and Relation Computation}
Given a language scene:
\begin{equation}
    \begin{aligned}
        \text{LS} &= \{V,\, O,\, E\}, \\
        V &= \{v_j\}_{j=1}^{L},\quad O = \{o_i\}_{i=1}^{N},\quad E = \{e_k\}_{k=1}^{M},
    \end{aligned}
    \label{eq:ls}
\end{equation}
where $V$ is the set of ego vehicles (AVs), $O$ is the set of surrounding vehicle objects, and $E$ is the set of road risk entities. Drawing on the spatial and topological relations that underpin human spatial cognition, we anchor on each $v_j$ and compute its relations to surrounding vehicles and to the risks on its road.

We first define the spatial direction relation $R_S$ between an AV and a surrounding vehicle. Given an OI record of an AV,
\begin{equation}
    v = [x_v,\, y_v,\, h_v,\, \text{ln}_v,\, \text{lx}_v,\, \text{rd}_v,\, \cdots]
    \label{eq:av_record}
\end{equation}
and an OI record of a surrounding vehicle,
\begin{equation}
    o = [x_o,\, y_o,\, h_o,\, \text{ln}_o,\, \text{lx}_o,\, \text{rd}_o,\, \cdots],
    \label{eq:obj_record}
\end{equation}
we transform the position of the target vehicle into a body frame aligned with the heading of the ego vehicle, which gives its longitudinal and lateral components:
\begin{equation}
    \begin{cases}
        \Delta x_{\text{lon}} = (x_o - x_v)\sin h_v + (y_o - y_v)\cos h_v \\[3pt]
        \Delta y_{\text{lat}} = (x_o - x_v)\cos h_v - (y_o - y_v)\sin h_v
    \end{cases}
    \label{eq:body_frame}
\end{equation}
From these components we compute the signed bearing angle $\theta \in (-\pi, \pi]$:
\begin{equation}
    \theta = \operatorname{atan2}\big(\Delta y_{\text{lat}},\, \Delta x_{\text{lon}}\big).
    \label{eq:theta}
\end{equation}
To keep the direction scheme free of any fixed granularity, we divide the space around the ego vehicle into $K$ uniform sectors, all referenced to its heading, and define the spatial direction relation between the AV and a surrounding vehicle as
\begin{equation}
    R_S^{(K)}(v_j, o_i) = \left\lfloor \frac{\theta + \pi/K}{2\pi/K} \right\rfloor \bmod K.
    \label{eq:spatial_relation}
\end{equation}
This paper sets $K = 4$, so the four values of $R_S$ correspond to front, left, rear, and right in order, which matches how drivers think about direction.

We also define lane topology relations from road identifiers and lane indices. Let $\Delta \text{lx} = \text{lx}_v - \text{lx}_o$ denote the difference in lane index. The $n$-th order lane neighborhood relation is
\begin{equation}
    R_L^{(n)} = \big\{ (v_j, o_i) \ \big|\ \text{rd}_v = \text{rd}_o \wedge |\Delta \text{lx}| \leq n \big\},
    \label{eq:lane_relation}
\end{equation}
where $\Delta \text{lx} = 0$ means the same lane, and $\Delta \text{lx} = 1$ and $\Delta \text{lx} = -1$ place the target vehicle in the lane to the left and to the right of the ego vehicle. This paper sets $n = 1$.

Beyond these vehicle-to-vehicle relations, we define relations between the ego vehicle and roads and road risks. We write the road that $v_j$ occupies as $\text{rd}(v_j)$ and define the vehicle-road relation as
\begin{equation}
    R_R = \{(v_j, \text{rd}) \mid \text{rd} = \text{rd}(v_j)\}.
    \label{eq:road_relation}
\end{equation}
Vehicle objects hang off the ego vehicle as their anchor, but road risk entities work differently. A risk entity $e_k$ rides on a road and belongs naturally to that road $\text{rd}(e_k)$, so a road name alone retrieves the risk information of any road. To capture how an ego vehicle relates to the risks around it, we define a $k$-th order road neighborhood on the road topology graph $\mathcal{G}$:
\begin{equation}
    \mathcal{N}_k(\text{rd}) = \{\text{rd}' \mid d_{\mathcal{G}}(\text{rd}, \text{rd}') \leq k \},
    \label{eq:road_neighborhood}
\end{equation}
where $d_{\mathcal{G}}(\cdot,\cdot)$ counts the hops along the shortest path in the road topology graph. With $R_R$ as the hub, we define the vehicle-risk relation $R_E$ as
\begin{equation}
    R_E^{(k)} = \big\{ (v_j, e_k) \ \big|\ \text{rd}(e_k) \in \mathcal{N}_k\big(\text{rd}(v_j)\big) \big\}.
    \label{eq:risk_relation}
\end{equation}
This paper sets $k = 1$, which links the ego vehicle to the risk entities on its own road and on adjacent roads. Putting these together, we build an AV-centric entity-relation graph:
\begin{equation}
    G = (\text{LS},\, R), \quad R = R_S \cup R_L \cup R_R \cup R_E.
    \label{eq:er_graph}
\end{equation}
Note that vehicle-to-vehicle relations carry an implicit spatial limit of 100 meters, while the road topology neighborhood $\mathcal{N}_k$ bounds the range of road risk association.

\subsubsection{Attribute-Entity-Relation Graph Construction}
An entity amounts to a set of attributes. When people ask a question, they usually pick out an entity with a few of its attributes and then ask about another attribute they do not know. Take the question ``What is the speed of the red truck slowing down on my right?'' The color, vehicle type, and acceleration form an attribute subset that, together with a relation ($R_S$: right), identifies the target vehicle. The question then asks for its unknown speed attribute. This observation motivates our attribute-entity-relation (AER) graph, which we describe below.

Given an LS, we attach attributes to the corresponding entities in $G$. Vehicle objects $o_i$ and ego vehicles $v_j$ carry object information (OI) attributes, and road risk entities $e_k$ carry environment information (EI) attributes. Table~\ref{tab:OI_EI_schema} gives the attribute schema of both. We then mask one attribute, written $a_{\text{masked}}$, as the query target and use the remaining attributes and relations to identify the entity. Formally, given a set of attribute constraints $\mathcal{A}$ and a relation $R$, we express the selection of target entities as
\begin{equation}
    \Phi(\mathcal{A}, R) = \Big\{ \xi \in O \cup E \ \Big|\ R(v_j, \xi) \wedge \bigwedge_{a \in \mathcal{A}} \big[ a(\xi) = a^{*} \big] \Big\}
    \label{eq:selection}
\end{equation}
where $\xi$ is a candidate entity, $a(\xi)$ is the value of attribute $a$ on entity $\xi$, and $a^{*}$ is the constraint value that the question gives for that attribute. The conjunction requires every attribute constraint to hold at once, and $\xi \in O \cup E$ shows that the same selection mechanism serves vehicle objects and road risk entities alike. The answer to the query is $a_{\text{masked}}(\xi)$. When $|\Phi(\mathcal{A}, R)| > 1$, the selection result also forms the basis of counting and existence queries. This enriches the entity-relation graph $G$ into an AER graph, shown in Fig.~\ref{fig_AER}. We build one AER graph for every AV in every LS.

\subsubsection{Question Template Design and Refinement}

\paragraph{Ego-anchored QA}
Within a given road space, drivers usually identify an entity of interest through its relation to the ego vehicle plus a subset of its attributes. They then ask about one unknown attribute, or about a value computed from attributes they already have. Following this pattern, we design five basic template types: attribute, distance, counting, existence, and safety-related metrics. We generate QA pairs through the mapping in Eq.~\eqref{eq:selection}. First, we pick a relation from $R$ and an attribute subset from the AER graph, then fill both into a randomly chosen question template. Second, we take the masked attribute, or a computed attribute such as distance or TTC, as the answer.

\paragraph{Road-anchored  QA}
Drivers also care about environment states that bear no direct relation to the ego vehicle, such as the condition of a particular road. Unlike ego-centric QA, these questions locate a vehicle entity or a road risk entity directly by road name, and the LS yields their answers directly.

We feed our hand-designed primitive templates into ChatGPT to diversify their wording. We then review the results by hand and keep 1{,}835 template variants. Table \ref{tab:Template} gives their distribution. We denote this template set as $T_Q$.

\subsubsection{QA Instantiation}
With $T_Q$ and the AER graphs, we can instantiate diverse QA pairs for every LS. To create SafeDM-QA, we repeatedly draw a question template at random from $T_Q$ for each LS and instantiate one QA pair with the AER graph. We continue until we obtain 23.5K QA pairs. Fig.~\ref{fig:dataset_stats} reports the statistics of SafeDM-QA.

\begin{figure*}
  \centering
  \includegraphics[width=\textwidth]{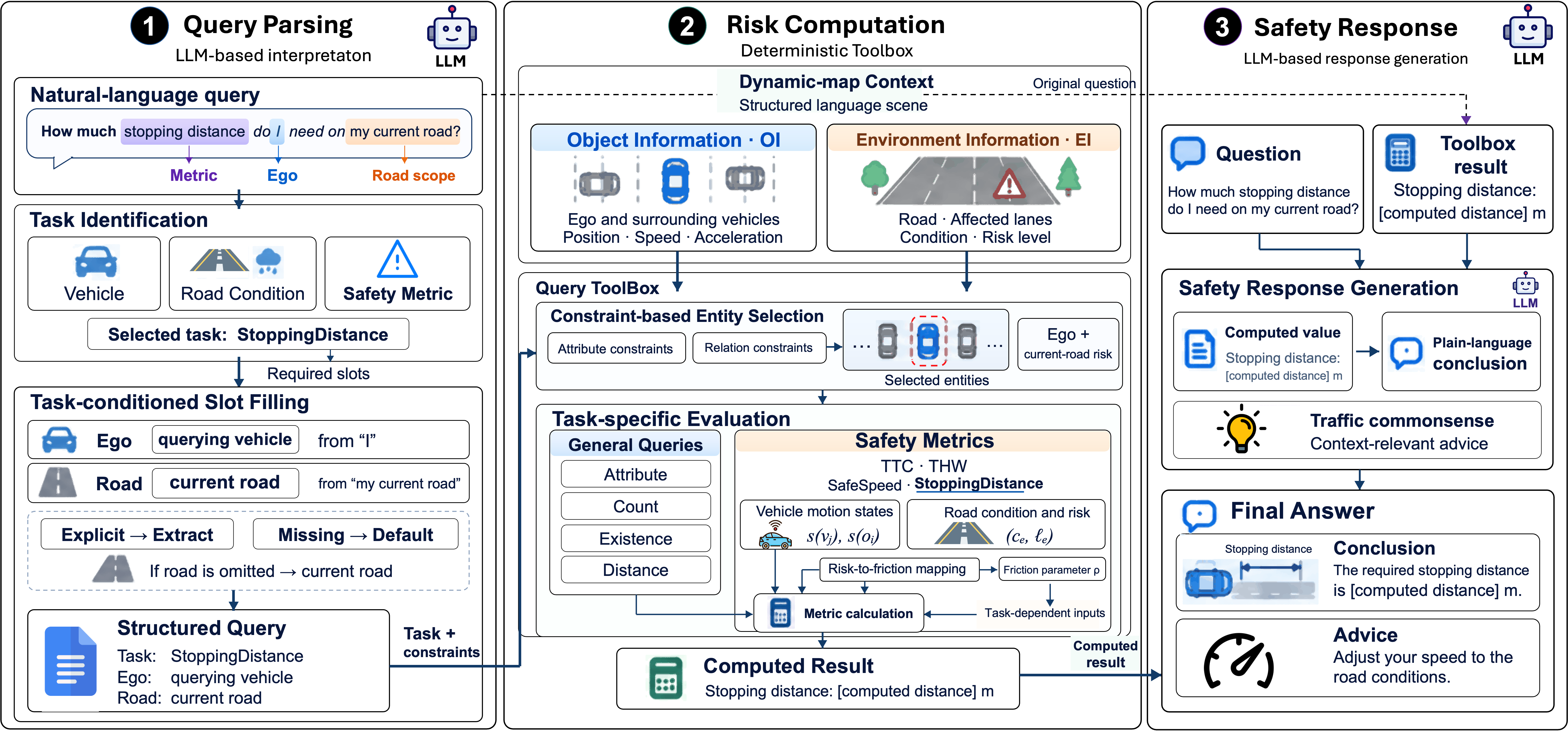}
  \caption{Workflow of SafeStep: (1) query parsing, (2) toolbox-based risk computation, and (3) safety response generation.}
  \label{fig_safestep}
\end{figure*}

\subsection{SafeStep Prompting Mechanism}

To support precise, safety-oriented natural language queries over the DM, we propose the Safety-aware Step-wise Prompting (SafeStep) mechanism. SafeStep breaks a query into three steps that build on one another: query parsing, risk computation, and safety response, as Fig.~\ref{fig_safestep} shows.

\subsubsection{Query Parsing}
Accurate querying starts with pinning down the query task behind the question. Given a question $Q$, the task parsing prompt $P_{tp}$ asks the LLM to analyze the question and output its single query task:
\begin{equation}
    \mathcal{T}_i = \text{LLM}(Q,\, P_{tp}), \quad \mathcal{T}_i \in \mathcal{T},
    \label{eq:task_parsing}
\end{equation}
where $\mathcal{T}$ is the predefined task set. It covers three kinds of query tasks: vehicle risks, road condition risks, and safety metrics that couple the two. Table~\ref{tab:tasks} gives the details.

Different tasks call for different query parameters. We define the parameter schema of task $\mathcal{T}_i$ as a slot set $\Psi(\mathcal{T}_i) = \{\psi_m\}_{m=1}^{M}$ and use the parameter parsing prompt $P_{pp}$ to have the LLM fill the slots one by one:
\begin{equation}
    \hat{p}_m =
    \begin{cases}
        \mathop{\arg\max}\limits_{p \in \mathbb{D}_m} P_{\theta}\left(p \mid Q,\, \psi_m,\, P_{pp}\right), & \psi_m \models Q,\\[4pt]
        \delta_m, & \text{otherwise},
    \end{cases}
    \label{eq:slot_parsing}
\end{equation}
where $\mathbb{D}_m$ is the value domain of slot $\psi_m$, the notation $\psi_m \models Q$ means that the question $Q$ mentions slot $\psi_m$ explicitly, and $\delta_m$ is the default value of that slot. The parsed slots together form the query parameters $\mathcal{P} = \{(\psi_m, \hat{p}_m)\}_{m=1}^{M} = \mathcal{P}_a \cup \mathcal{P}_r$, where $\mathcal{P}_a$ holds the attribute constraints and $\mathcal{P}_r$ the relation constraints. We set each default $\delta_m$ to whatever matters most for the safety of the ego vehicle: the relation slot defaults to \textit{surrounding} for vehicle and safety metric queries, and to the road that the ego vehicle currently occupies for road condition queries.

\subsubsection{Risk Computation}
Given the query task $\mathcal{T}_i$ and the parameters $\mathcal{P}$, the toolbox runs a deterministic computation over the language scene LS. It first uses $\mathcal{P}$ to select the query objects that satisfy every constraint at once:

\begin{equation}
\begin{aligned}
\mathcal{C} = \biggl\{\, x \in \mathcal{A}_{\mathcal{T}_i} \;\bigg|\;
  &\bigwedge_{(R,\,r) \in \mathcal{P}_r} \big[R(v_j, x) = r\big] \\
  \wedge &\bigwedge_{(a,\,\alpha) \in \mathcal{P}_a} \big[a(x) = \alpha\big] \biggr\}
\end{aligned}
\label{eq:constraint_filter}
\end{equation}

where $v_j$ is the ego vehicle and $\mathcal{A}_{\mathcal{T}_i}$ is the entity domain of task $\mathcal{T}_i$, that is, either the vehicle set $O$ or the road condition entity set $E$. The relation $R$ comes from the relation scheme of Section~\ref{SafeDM-QA}, and $r$ is its value. The term $a(x)$ is an attribute of entity $x$, and $\alpha$ is its value. The toolbox then applies an evaluation operator $\Omega_{\mathcal{T}_i}$ to $\mathcal{C}$ according to the task type, which yields the numeric result:
\begin{equation}
    \mathcal{R}_n = \text{ToolBox}(\mathcal{T}_i, \mathcal{P}, \text{LS}) = \Omega_{\mathcal{T}_i}(\mathcal{C})
    \label{eq:toolbox}
\end{equation}
where $\Omega_{\mathcal{T}_i}$ handles counting, existence checking, attribute reading, or distance computation, depending on the task type. The toolbox thus amounts to a composite operator that filters by constraints and then evaluates.

For safety metric tasks, the evaluation operator goes further and couples vehicle risks with road condition risks:
\begin{equation}
    \mathcal{I}_\tau = \mathcal{F}_\tau\left(s(v_j),\, s(o_i),\, \rho\right), \quad
    \rho = \min\Big(\rho_0,\ \min_{e \in E(v_j)} \rho\left(c_e,\, \ell_e\right)\Big)
    \label{eq:safety_metric}
\end{equation}
where $\tau \in \{\text{TTC}, \text{THW}, \text{SafeSpeed}, \text{StoppingDistance}\}$. The terms $s(v_j)$ and $s(o_i)$ give the motion states of the ego vehicle and the target vehicle. The coefficient $\rho$ is the road friction coefficient and $\rho_0$ its value on a dry surface. The set $E(v_j)$ holds the road condition entities on the road of the ego vehicle, and $(c_e, \ell_e)$ are the road condition category and risk level of entity $e$, which the road condition risk assessment VLM outputs.

\subsubsection{Safety Response}
The numeric result from the toolbox still has to become an answer that serves driving safety. Through the safety response prompt $P_{sr}$, we have the LLM combine the computed result with traffic commonsense and generate the response:
\begin{equation}
    (\mathcal{R}_s,\, \mathcal{V}) = \text{LLM}(Q,\, \mathcal{R}_n,\, P_{sr}), \quad A = \mathcal{R}_s \oplus \mathcal{V},
    \label{eq:safety_response}
\end{equation}
where $\mathcal{R}_s$ is the semantic query conclusion, $\mathcal{V}$ is the driving advice, and $A$ is the final answer to the question $Q$.

Putting the three steps together, we formalize the natural language safety query of SafeStep as:
\begin{equation}
    A = \text{LLM}\Big(Q,\ \text{ToolBox}\big(\mathcal{T}_i,\ \text{LLM}(Q, \Psi(\mathcal{T}_i), P_{pp}),\ \text{LS}\big),\ P_{sr}\Big)
    \label{eq:safestep_full}
\end{equation}

% \begin{table}[t]
% \centering
% \caption{Schema of environment information}
% \label{tab:ei}
% \begin{tabular}{ll}
% \toprule
% \textbf{Field} & \textbf{Description} \\
% \midrule
% id & identifier \\
% rd & the current road \\
% rc & risk category\\
% rl & risk level (0--3) \\
% af & affected lane \\
% ds & the detecting sensor \\
% t  & time stamp \\
% \bottomrule
% \end{tabular}
% \end{table}

\section{Experiments}

Our experiments have two parts: evaluating Safety4DM on the risk perception task, and evaluating it on the natural language spatial (NLS) query task.

\subsection{Experimental Setup}

% \begin{table}[t]
% \centering
% \caption{Overview of LLM Configurations.}
% \label{tab:llm}
% \begin{tabular}{llcccc}
% \hline\hline
% \multicolumn{2}{c}{\multirow{2}{*}{\textbf{Model}}} & \textbf{Archi-} & \textbf{Context} & \textbf{Embedding} & \textbf{Quanti-} \\
% \multicolumn{2}{c}{} & \textbf{tecture} & \textbf{length} & \textbf{length} & \textbf{zation} \\
% \cline{1-2}
% \textbf{Family} & \textbf{Size} & & & & \\
% \hline
% GPT-oss & 20B & gptoss & 131072 & 2880 & MXFP4 \\
% \hline
% & 1.5B & qwen2 & 131072 & 1536 & Q4\_K\_M \\
% Deepseek-r1 & 7B & qwen2 & 131072 & 4096 & Q4\_K\_M \\
% & 14B & qwen2 & 131072 & 5120 & Q4\_K\_M \\
% & 32B & qwen2 & 131072 & 5120 & Q4\_K\_M \\
% \hline
% & 0.6B & qwen3 & 40960 & 1024 & Q4\_K\_M \\
% & 1.7B & qwen3 & 40960 & 2048 & Q4\_K\_M \\
% Qwen3 & 4B & qwen3 & 262144 & 2560 & Q4\_K\_M \\
% & 8B* & qwen3 & 40960 & 4096 & Q4\_K\_M \\
% & 14B & qwen3 & 40960 & 5120 & Q4\_K\_M \\
% & 30B & qwen3moe & 40960 & 5120 & Q4\_K\_M \\
% \hline
% Magistral & 24B & llama & 40000 & 5120 & Q4\_K\_M \\
% \hline
% & 1B & gemma3 & 32768 & 1152 & Q4\_K\_M \\
% Gemma & 4B & gemma3 & 131072 & 2560 & Q4\_K\_M \\
% & 12B & gemma3 & 131072 & 3840 & Q4\_K\_M \\
% & 27B & gemma3 & 131072 & 5376 & Q4\_K\_M \\
% \hline
% Llama3 & 8B & llama & 8192 & 4096 & Q4\_0 \\
% \hline\hline
% \multicolumn{6}{l}{\footnotesize *Reference model used in Safety4DM development.} \\
% \end{tabular}
% \end{table}

\paragraph{Datasets.}
We randomly split the RoadRisk dataset into training and test sets at a ratio of 4:1, and use the test set to evaluate the road safety risk perception LLM. We also randomly sample 7.3K question--answer pairs from SafeDM-QA to evaluate the NLS query performance of Safety4DM. These samples cover all query types (see Table \ref{tab:qa_count_test}) and span 400 VRC collaboration scenarios, with an even spread across the full simulation lifecycle.

\paragraph{Baselines.}
For the risk perception task, we use the 2B, 4B, and 8B models of the Qwen3-VL series as base models and compare their performance before and after SFT. This comparison shows whether domain-specific fine-tuning is necessary and how base model size affects performance. For the NLS query task, we compare 17 reasoning LLMs from six model families. The SafeStep mechanism in Safety4DM does not depend on any particular model and works with different LLMs out of the box, so this experiment tests how well it generalizes across models.

\paragraph{Evaluation Metrics.}
Both tasks share a two-stage structure: the model first determines a category and then produces a final result based on that category. We therefore use a unified set of metrics. Let $\mathcal{X}$ denote the first-stage target, which is the road condition category in the risk perception task and the query type in the query task. We define the classification accuracy $\mathcal{A}_C$ and the final accuracy $\mathcal{A}_F$ as:
\begin{equation}
    \mathcal{A}_C = \frac{\sum_{i=1}^{N} \mathbb{I}\left(\hat{\mathcal{X}}_i = \mathcal{X}_i\right)}{N} \times 100\%, \quad
    \mathcal{A}_F = \frac{N_c}{N} \times 100\%,
    \label{eq:accuracy}
\end{equation}
where $N$ is the total number of samples, $\mathbb{I}(\cdot)$ is the indicator function, and $N_c$ is the number of samples with a fully correct final result. In the risk perception task, a correct sample needs both the right road condition category and the right risk level. In the query task, the query result must exactly match the ground truth.

To measure how the performance of a model $\mathcal{M}_k$ varies across categories or tasks, let $\mathcal{G}_i$ denote the category or task under study. We define the bias indicator as:
\begin{equation}
    \mathcal{B}_{\mathcal{G}_i}^{\mathcal{M}_k} =
    \begin{cases}
        +1, & \text{if } \mathcal{A}_F^{\mathcal{M}_k}(\mathcal{G}_i) \geq \mathcal{A}_F^{\mathcal{M}_k}(\mathcal{G}_j),\ \forall j \neq i; \\
        -1, & \text{if } \mathcal{A}_F^{\mathcal{M}_k}(\mathcal{G}_i) < \mathcal{A}_F^{\mathcal{M}_k}(\mathcal{G}_j),\ \forall j \neq i; \\
        0,  & \text{otherwise.}
    \end{cases}
    \label{eq:bias_indicator}
\end{equation}
Summing this indicator over all models gives the performance bias of each category or task:
\begin{equation}
    \mathcal{B}_{\mathcal{G}_i} = \sum_{k} \mathcal{B}_{\mathcal{G}_i}^{\mathcal{M}_k}.
    \label{eq:bias}
\end{equation}
In the rest of this paper, $\mathcal{A}_P$ and $\mathcal{A}_Q$ denote the final accuracy $\mathcal{A}_F$ on the risk perception task and the query task, respectively.

\paragraph{Implementation Details.}
We fine-tune the road safety risk perception LLM with SFT in the LLaMA-Factory framework, using LoRA for parameter-efficient training. We set the LoRA rank to 32 and the learning rate to $1.0 \times 10^{-4}$. We use a per-device batch size of 8 with 2 gradient accumulation steps and train for 2 epochs. To plug different LLMs into the NLS query task, we use the same set of SafeStep prompts $\{P_{tp}, P_{pp}, P_{sr}\}$ for every model. We developed and refined these prompts with Qwen3:8B as the reference model, then applied them to all other models without changes. This setup lets us test how well Safety4DM generalizes. We run all experiments locally on a single RTX 4080 GPU with 32\,GB of memory.

% Packages: \usepackage{booktabs, multirow, makecell}
\begin{table}[t]
\centering
\caption{Question Count Distribution of the Evaluation Subset.}
\label{tab:qa_count_test}
\footnotesize
\setlength{\tabcolsep}{2pt}
\renewcommand{\arraystretch}{1.1}
\begin{tabular}{@{}l l c r r r@{}}
\toprule
\textbf{Query type} & \textbf{Task} & \textbf{Domain} &
\makecell{\textbf{Ego-}\\\textbf{anchored}} &
\makecell{\textbf{Road-}\\\textbf{anchored}} &
\textbf{Total} \\
\midrule
\multirow{9}{*}{attribute query}
  & velocity         & OI    & 280 & 24  & 304 \\
  & acceleration     & OI    & 300 & 19  & 319 \\
  & heading          & OI    & 292 & 19  & 311 \\
  & color            & OI    & 278 & 18  & 296 \\
  & classification   & OI    & 275 & 22  & 297 \\
  & size             & OI    & 276 & 24  & 300 \\
  & status           & OI    & 277 & 22  & 299 \\
  & condition type   & EI    & 400 & 237 & 637 \\
  & risk level       & EI    & 163 & 237 & 400 \\
\midrule
\multirow{4}{*}{safety query}
  & TTC               & OI+EI & 268 & 81  & 349 \\
  & THW               & OI+EI & 257 & 81  & 338 \\
  & stopping distance & OI+EI & 238 & 117 & 355 \\
  & safe speed        & OI+EI & 278 & 101 & 379 \\
\midrule
\multicolumn{2}{@{}l}{distance query}  & OI    & 274 & 19  & 293  \\
\multicolumn{2}{@{}l}{count query}     & OI/EI & 498 & 801 & 1299 \\
\multicolumn{2}{@{}l}{existence query} & OI/EI & 781 & 309 & 1090 \\
\midrule
\multicolumn{3}{@{}l}{\textbf{Total}} & \textbf{5135} & \textbf{2131} & \textbf{7266} \\
\bottomrule
\end{tabular}
\end{table}

\begin{table}
\renewcommand\arraystretch{1.3}
\centering
\caption{Zero-Shot Results of the Road Risk Perception Task.}
\label{tab:risk_zs}
\setlength{\tabcolsep}{4pt}
\begin{tabular}{lccccccc}
\hline\hline
\textbf{Model} & \multicolumn{2}{c}{\textbf{Qwen3-VL-2B}} & \multicolumn{2}{c}{\textbf{Qwen3-VL-4B}} & \multicolumn{2}{c}{\textbf{Qwen3-VL-8B}} & \multirow{2}{*}{$\mathcal{B}_{c}$} \\
\cline{2-7}
\textbf{Road condition} & $\mathcal{A}_C$ & $\mathcal{A}_P$ & $\mathcal{A}_C$ & $\mathcal{A}_P$ & $\mathcal{A}_C$ & $\mathcal{A}_P$ & \\
\hline
Normal     & 38.33 & \textbf{38.33} & 91.67 & \textbf{91.67} & 93.00 & \textbf{93.00} & +3 \\
Floodwater & 82.33 & 29.67 & 96.33 & 58.67 & 97.67 & 69.67 & 0 \\
Snow       & 86.33 & 35.00 & 92.67 & 64.67 & 90.33 & \underline{56.00} & -1 \\
Obstacle   & 48.67 & \underline{27.00} & 61.67 & \underline{49.00} & 70.67 & 62.67 & -2 \\
\hline
\textbf{Average} & 63.92 & 32.50 & 85.58 & 66.00 & \textbf{87.92} & \textbf{70.33} & -- \\
\hline\hline
\end{tabular}
\end{table}

\begin{table}
\renewcommand\arraystretch{1.3}
\centering
\caption{Results of the Road Risk Perception Task after SFT.}
\label{tab:risk_sft}
\setlength{\tabcolsep}{4pt}
\begin{tabular}{lccccccc}
\hline\hline
\textbf{Model} & \multicolumn{2}{c}{\textbf{Qwen3-VL-2B}} & \multicolumn{2}{c}{\textbf{Qwen3-VL-4B}} & \multicolumn{2}{c}{\textbf{Qwen3-VL-8B}} & \multirow{2}{*}{$\mathcal{B}_{c}$} \\
\cline{2-7}
\textbf{Road condition} & $\mathcal{A}_C$ & $\mathcal{A}_P$ & $\mathcal{A}_C$ & $\mathcal{A}_P$ & $\mathcal{A}_C$ & $\mathcal{A}_P$ & \\
\hline
Normal     & 97.67 & \textbf{97.67} & 96.00 & \textbf{96.00} & 97.33 & \textbf{97.33} & +3 \\
Floodwater & 99.33 & \underline{77.33} & 99.00 & \underline{78.67} & 99.33 & \underline{78.00} & -3 \\
Snow       & 99.33 & 88.33 & 99.33 & 93.00 & 97.67 & 91.33 & 0 \\
Obstacle   & 100 & 87.67 & 100 & 91.67 & 100 & 91.67 & 0 \\
\hline
\textbf{Average} & 99.08 & 87.75 & 98.58 & \textbf{89.83} & 98.58 & 89.58 & -- \\
\hline\hline
\end{tabular}
\end{table}

\subsection{Risk Perception Task}

Table~\ref{tab:risk_zs} reports the zero-shot results of the Qwen3-VL models on the risk perception task without fine-tuning. Overall, performance improves as model size grows. The 2B model performs worst on both road condition classification and risk grading. It also produces many format parsing errors (233 cases), which shows that it struggles to follow the required structured output format. The 4B model makes only a handful of parsing errors (2 cases) and improves on both classification and grading. The 8B model performs better still, reaching 87.92\% in $\mathcal{A}_C$ and 70.33\% in $\mathcal{A}_P$. However, even the larger 8B model does not classify road conditions or grade risks well enough for the risk perception task in Safety4DM.

We then fine-tune these models with SFT. Table~\ref{tab:risk_sft} shows the results. Fine-tuning greatly improves the overall performance of every model and removes all format parsing errors. After fine-tuning, the 2B and 8B models differ by only 1.83 percentage points in $\mathcal{A}_P$, even though the 8B model has four times as many parameters. This shows that smaller models gain the most from fine-tuning. The 2B model improves its $\mathcal{A}_P$ by 55.25 percentage points, far more than the 4B model (23.83) and the 8B model (19.25). On this task, domain-specific fine-tuning brings much larger gains than scaling up the model. In terms of category bias, the fine-tuned models perform poorly when grading standing water ($\mathcal{B}_c = -3$). Specifically, the fine-tuned models reach 99.00\%--99.33\% in $\mathcal{A}_C$ for standing water, but only 77.33\%--78.67\% in $\mathcal{A}_P$. Detecting water on the road is easy; the hard part is grading its depth. Even after fine-tuning, the models still grade this road condition with limited accuracy.

\begin{table*}[t]
\renewcommand\arraystretch{1.3}
\setlength{\tabcolsep}{2.5mm}
\centering
\caption{Experimental Results of Safety4DM across Different LLM Families.}
\label{tab:family}
\begin{tabular}{lccccccccccccc}
\hline\hline
\textbf{Safety4DM using:} & \multicolumn{2}{c}{\textbf{DS-r1:32B}} & \multicolumn{2}{c}{\textbf{Qwen3:30B}} & \multicolumn{2}{c}{\textbf{GPT-oss:20B}} & \multicolumn{2}{c}{\textbf{Magistral:24B}} & \multicolumn{2}{c}{\textbf{Gemma3:27B}} & \multicolumn{2}{c}{\textbf{Llama3.1:8B}} & \multirow{2}{*}{$\mathcal{B}_{\mathcal{T}_i}$} \\
\cline{2-13}
\textbf{Query task} & $\mathcal{A}_C$ & $\mathcal{A}_Q$ & $\mathcal{A}_C$ & $\mathcal{A}_Q$ & $\mathcal{A}_C$ & $\mathcal{A}_Q$ & $\mathcal{A}_C$ & $\mathcal{A}_Q$ & $\mathcal{A}_C$ & $\mathcal{A}_Q$ & $\mathcal{A}_C$ & $\mathcal{A}_Q$ & \\
\hline
(1) ConditionType  & 100 & \textbf{100} & 100 & \textbf{100} & 100 & \textbf{100} & 99.37 & 97.33 & 100 & 98.90 & 44.74 & 44.74 & +3 \\
(2) RiskLevel      & 100 & \textbf{100} & 100 & 99.25 & 100 & \textbf{100} & 98.25 & 98.25 & 100 & 99.00 & 44.25 & 44.25 & +2 \\
(3) EI Count       & 99.29 & 97.98 & 99.19 & 97.78 & 100 & 98.28 & 100 & \textbf{98.28} & 99.09 & 97.07 & 85.37 & 72.25 & +1 \\
(4) EI Exist       & 100 & \textbf{100} & 96.98 & 97.89 & 100 & \textbf{100} & 94.86 & 94.86 & 99.09 & \textbf{99.09} & 34.44 & \underline{26.28} & +2 \\
(5) Velocity       & 100 & 96.71 & 100 & 97.37 & 98.68 & 97.04 & 100 & 90.13 & 100 & 95.72 & 100 & \textbf{91.12} & +1 \\
(6) Acceleration   & 97.49 & 91.22 & 93.73 & 88.09 & 91.22 & 88.09 & 96.24 & 87.46 & 98.43 & 86.52 & 100 & 84.64 & 0 \\
(7) Heading        & 93.57 & 91.64 & 93.57 & \underline{87.46} & 91.96 & 87.46 & 88.10 & \underline{76.53} & 94.21 & 83.92 & 87.46 & 63.99 & -2 \\
(8) Color          & 100 & 92.91 & 100 & 91.22 & 96.96 & 91.22 & 100 & 95.61 & 100 & 88.18 & 100 & 80.74 & 0 \\
(9) Classification & 100 & 92.59 & 100 & 92.26 & 98.99 & 90.91 & 100 & 88.89 & 100 & 81.14 & 99.33 & 86.53 & 0 \\
(10) Size          & 100 & 98.00 & 100 & 96.00 & 99.00 & 96.33 & 99.67 & 95.00 & 100 & 92.00 & 70.67 & 62.67 & 0 \\
(11) Status        & 88.63 & \underline{87.96} & 91.97 & 89.97 & 87.63 & \underline{86.96} & 100 & 95.99 & 87.96 & \underline{79.60} & 87.96 & 73.58 & -3 \\
(12) Distance      & 100 & 96.25 & 100 & 95.90 & 99.66 & 94.54 & 100 & 94.88 & 100 & 94.54 & 100 & 88.74 & 0 \\
(13) OI Count      & 100 & 98.70 & 100 & 98.70 & 99.03 & 97.73 & 100 & 96.75 & 100 & 95.45 & 92.86 & 86.69 & 0 \\
(14) OI Exist      & 100 & 99.07 & 100 & 97.66 & 97.90 & 94.63 & 100 & 95.56 & 92.76 & 91.36 & 94.86 & 88.08 & 0 \\
(15) TTC           & 97.99 & 98.28 & 100 & \textbf{100} & 99.43 & 99.43 & 98.28 & 96.28 & 93.12 & 92.55 & 100 & 82.52 & +1 \\
(16) THW           & 100 & 97.93 & 100 & 98.52 & 99.11 & 99.11 & 100 & 94.67 & 100 & 97.34 & 96.15 & 73.96 & 0 \\
(17) SafeSpeed     & 100 & \textbf{100} & 100 & \textbf{100} & 98.15 & 97.89 & 100 & 97.36 & 100 & 98.94 & 100 & 66.75 & +2 \\
(18) StoppingDist. & 100 & \textbf{100} & 100 & 99.44 & 98.87 & 98.59 & 100 & 97.75 & 100 & 97.75 & 100 & 74.93 & +1 \\
\hline
\textbf{Average}   & \textbf{98.95} & \textbf{97.23} & 98.73 & 96.59 & 98.10 & 96.28 & 98.61 & 94.72 & 98.22 & 93.92 & 81.06 & 68.30 & -- \\
\hline\hline
\end{tabular}
\end{table*}

\begin{table*}[h]
\renewcommand\arraystretch{1.3}
\setlength{\tabcolsep}{2.5mm}
\centering
\caption{Experimental Results of Safety4DM using the Qwen3 Family.}
\label{tab:qwen}
\begin{tabular}{lccccccccccccc}
\hline\hline
\textbf{Safety4DM using} & \multicolumn{2}{c}{\textbf{Qwen3:0.6B}} & \multicolumn{2}{c}{\textbf{Qwen3:1.7B}} & \multicolumn{2}{c}{\textbf{Qwen3:4B}} & \multicolumn{2}{c}{\textbf{Qwen3:8B}} & \multicolumn{2}{c}{\textbf{Qwen3:14B}} & \multicolumn{2}{c}{\textbf{Qwen3:30B}} & \multirow{2}{*}{$\mathcal{B}_{\mathcal{T}_i}$} \\
\cline{2-13}
\textbf{Query task} & $\mathcal{A}_C$ & $\mathcal{A}_Q$ & $\mathcal{A}_C$ & $\mathcal{A}_Q$ & $\mathcal{A}_C$ & $\mathcal{A}_Q$ & $\mathcal{A}_C$ & $\mathcal{A}_Q$ & $\mathcal{A}_C$ & $\mathcal{A}_Q$ & $\mathcal{A}_C$ & $\mathcal{A}_Q$ & \\
\hline
(1) ConditionType  & 69.86 & 61.70 & 70.80 & 68.45 & 100 & \textbf{99.84} & 100 & \textbf{99.69} & 100 & 99.84 & 100 & \textbf{100} & +3 \\
(2) RiskLevel      & 74.75 & 66.25 & 76.50 & 70.75 & 100 & 95.50 & 96.50 & 92.00 & 100 & \textbf{100} & 100 & 99.25 & +1 \\
(3) EI Count       & 85.07 & 67.51 & 88.90 & 87.08 & 100 & 98.28 & 100 & 98.28 & 100 & 98.28 & 99.19 & 97.78 & 0 \\
(4) EI Exist       & 53.17 & 40.03 & 26.89 & \underline{26.59} & 96.22 & 96.07 & 96.98 & 95.17 & 100 & \textbf{100} & 96.98 & 97.89 & 0 \\
(5) Velocity       & 99.01 & 43.09 & 98.68 & 79.28 & 100 & 97.04 & 100 & 92.76 & 100 & 96.05 & 100 & 97.37 & 0 \\
(6) Acceleration   & 54.55 & 28.21 & 61.44 & 51.72 & 99.06 & 92.79 & 97.18 & 91.22 & 97.18 & 91.54 & 93.73 & 88.09 & 0 \\
(7) Heading        & 82.64 & 30.87 & 88.10 & 55.31 & 83.60 & \underline{76.85} & 84.57 & \underline{79.74} & 89.07 & 84.89 & 93.57 & \underline{87.46} & -3 \\
(8) Color          & 93.92 & \underline{11.82} & 95.27 & 31.08 & 100 & 81.08 & 100 & 80.74 & 100 & 90.20 & 100 & 91.22 & -1 \\
(9) Classification & 76.09 & 22.22 & 93.94 & 67.34 & 100 & 88.89 & 100 & 91.25 & 98.99 & 90.57 & 100 & 92.26 & 0 \\
(10) Size          & 83.00 & 31.00 & 91.33 & 57.33 & 99.67 & 96.67 & 99.67 & 96.00 & 99.33 & 96.33 & 100 & 96.00 & 0 \\
(11) Status        & 27.42 & 14.72 & 64.55 & 52.17 & 87.96 & 87.63 & 84.95 & 83.28 & 82.94 & \underline{82.61} & 91.97 & 89.97 & -1 \\
(12) Distance      & 97.95 & 58.02 & 99.66 & 71.33 & 99.66 & 93.52 & 99.66 & 94.20 & 99.66 & 94.88 & 100 & 95.90 & 0 \\
(13) OI Count      & 98.70 & 58.77 & 98.70 & 76.30 & 100 & 98.70 & 98.38 & 97.73 & 99.35 & 98.70 & 100 & 98.70 & 0 \\
(14) OI Exist      & 93.46 & \textbf{74.77} & 99.77 & \textbf{94.63} & 100 & 97.43 & 99.07 & 95.79 & 99.30 & 96.96 & 100 & 97.66 & +2 \\
(15) TTC           & 95.42 & 26.36 & 97.13 & 73.07 & 100 & 97.71 & 98.85 & 99.14 & 100 & 98.28 & 100 & \textbf{100} & +1 \\
(16) THW           & 99.11 & 21.01 & 98.52 & 70.12 & 100 & 93.49 & 100 & 97.63 & 100 & 98.52 & 100 & 98.52 & 0 \\
(17) SafeSpeed     & 95.25 & 21.64 & 98.94 & 68.60 & 100 & 98.68 & 100 & 99.47 & 100 & \textbf{100} & 100 & \textbf{100} & +2 \\
(18) StoppingDist. & 96.90 & 30.70 & 98.59 & 73.24 & 100 & 95.21 & 100 & 98.03 & 100 & 99.72 & 100 & 99.44 & 0 \\
\hline
\textbf{Average}   & 80.79 & 43.66 & 83.04 & 66.30 & 98.39 & 94.66 & 97.92 & 94.45 & 98.55 & 96.31 & \textbf{98.73} & \textbf{96.59} & -- \\
\hline\hline
\end{tabular}
\end{table*}

\begin{table*}[h]
\renewcommand\arraystretch{1.6}
\centering
\caption{Experimental Results of Safety4DM using the Gemma3 and DeepSeek-r1 Families.}
\label{tab:gemma_deepseek}
% NOTE: Gemma3:27B and DS-r1:32B columns hold PREDICTED 7.3k values (folded to integer counts); replace with measured results.
\resizebox{\textwidth}{!}{
\begin{tabular}{l*{9}{c}@{\hspace{5pt}}c@{\hspace{5pt}}*{9}{c}}
\hline\hline
\multirow{3}{*}{\makecell[l]{\textbf{Safety4DM using}\\[2pt]\textbf{Query task}}}
& \multicolumn{9}{c}{\textbf{Gemma3 family}} & & \multicolumn{9}{c}{\textbf{DeepSeek-r1 family}} \\
\cline{2-10}\cline{12-20}
& \multicolumn{2}{c}{\textbf{Gemma3:1B}} & \multicolumn{2}{c}{\textbf{Gemma3:4B}} & \multicolumn{2}{c}{\textbf{Gemma3:12B}} & \multicolumn{2}{c}{\textbf{Gemma3:27B}} & \multirow{2}{*}{$\mathcal{B}_{\mathcal{T}_i}$}
& & \multicolumn{2}{c}{\textbf{DS-r1:1.5B}} & \multicolumn{2}{c}{\textbf{DS-r1:7B}} & \multicolumn{2}{c}{\textbf{DS-r1:14B}} & \multicolumn{2}{c}{\textbf{DS-r1:32B}} & \multirow{2}{*}{$\mathcal{B}_{\mathcal{T}_i}$} \\
\cline{2-9}\cline{12-19}
& $\mathcal{A}_C$ & $\mathcal{A}_Q$ & $\mathcal{A}_C$ & $\mathcal{A}_Q$ & $\mathcal{A}_C$ & $\mathcal{A}_Q$ & $\mathcal{A}_C$ & $\mathcal{A}_Q$ & &
& $\mathcal{A}_C$ & $\mathcal{A}_Q$ & $\mathcal{A}_C$ & $\mathcal{A}_Q$ & $\mathcal{A}_C$ & $\mathcal{A}_Q$ & $\mathcal{A}_C$ & $\mathcal{A}_Q$ & \\
\hline
(1) ConditionType & 16.64 & \underline{4.08} & 100 & \textbf{100} & 100 & \textbf{100} & 100 & 98.90 & +1 &  & 30.30 & 18.52 & 20.41 & \underline{16.01} & 100 & \textbf{100} & 100 & \textbf{100} & +1 \\
(2) RiskLevel & 60.75 & 16.50 & 98.25 & 98.25 & 100 & \textbf{100} & 100 & 99.00 & +1 &  & 29.25 & 24.25 & 39.25 & 35.75 & 100 & \textbf{100} & 100 & \textbf{100} & +2 \\
(3) EI Count & 0.00 & 9.18 & 53.68 & \underline{50.55} & 65.09 & \underline{64.18} & 99.09 & 97.07 & -2 &  & 6.46 & \underline{4.34} & 70.13 & 59.23 & 100 & 97.07 & 99.29 & 97.98 & -1 \\
(4) EI Exist & 0.91 & 13.44 & 91.69 & 91.54 & 98.19 & 98.19 & 99.09 & \textbf{99.09} & +1 &  & 12.54 & 6.50 & 34.89 & 20.39 & 99.70 & 99.70 & 100 & \textbf{100} & +1 \\
(5) Velocity & 77.96 & 60.86 & 100 & 81.58 & 100 & 98.36 & 100 & 95.72 & 0 &  & 68.42 & 24.67 & 100 & 73.36 & 100 & 93.75 & 100 & 96.71 & 0 \\
(6) Acceleration & 34.80 & 26.65 & 89.66 & 78.37 & 87.77 & 77.43 & 98.43 & 86.52 & 0 &  & 52.98 & 13.79 & 69.59 & 42.95 & 92.48 & 86.83 & 97.49 & 91.22 & 0 \\
(7) Heading & 39.55 & 23.15 & 90.68 & 67.85 & 99.04 & 88.10 & 94.21 & 83.92 & 0 &  & 37.62 & 9.65 & 81.67 & 42.44 & 90.03 & 83.28 & 93.57 & 91.64 & 0 \\
(8) Color & 100 & 44.59 & 100 & 88.85 & 100 & 91.55 & 100 & 88.18 & 0 &  & 71.62 & 30.07 & 99.66 & 34.80 & 100 & 79.39 & 100 & 92.91 & 0 \\
(9) Classification & 49.83 & 23.23 & 100 & 85.19 & 99.66 & 92.26 & 100 & 81.14 & 0 &  & 24.92 & 9.43 & 93.60 & 48.48 & 100 & 91.58 & 100 & 92.59 & 0 \\
(10) Size & 42.67 & 31.00 & 100 & 78.67 & 100 & 91.33 & 100 & 92.00 & 0 &  & 43.00 & 19.33 & 99.67 & 69.00 & 100 & 97.00 & 100 & 98.00 & 0 \\
(11) Status & 64.21 & 28.43 & 87.96 & 79.60 & 87.96 & 81.27 & 87.96 & \underline{79.60} & -1 &  & 19.73 & 5.35 & 61.87 & 32.11 & 80.27 & \underline{79.26} & 88.63 & \underline{87.96} & -2 \\
(12) Distance & 100 & \textbf{67.24} & 100 & 93.17 & 100 & 96.25 & 100 & 94.54 & +1 &  & 59.39 & 25.60 & 99.32 & 65.19 & 100 & 94.20 & 100 & 96.25 & 0 \\
(13) OI Count & 100 & 59.09 & 100 & 90.26 & 99.35 & 95.78 & 100 & 95.45 & 0 &  & 83.77 & 38.31 & 98.38 & 69.48 & 99.35 & 98.05 & 100 & 98.70 & 0 \\
(14) OI Exist & 11.21 & 32.71 & 100 & 90.89 & 96.03 & 95.09 & 92.76 & 91.36 & 0 &  & 68.69 & \textbf{49.30} & 95.79 & \textbf{82.71} & 100 & 98.36 & 100 & 99.07 & +2 \\
(15) TTC & 75.07 & 34.67 & 92.55 & 80.52 & 91.98 & 93.70 & 93.12 & 92.55 & 0 &  & 82.23 & 6.59 & 97.99 & 43.55 & 99.43 & 97.71 & 97.99 & 98.28 & 0 \\
(16) THW & 87.87 & 32.54 & 100 & 84.32 & 100 & 90.83 & 100 & 97.34 & 0 &  & 69.23 & 10.06 & 96.75 & 38.46 & 100 & 97.34 & 100 & 97.93 & 0 \\
(17) SafeSpeed & 20.58 & 4.49 & 100 & 83.11 & 100 & 95.25 & 100 & 98.94 & 0 &  & 75.46 & 12.14 & 100 & 49.60 & 100 & 98.42 & 100 & \textbf{100} & +1 \\
(18) StoppingDist. & 35.21 & 9.01 & 100 & 81.13 & 100 & 88.73 & 100 & 97.75 & 0 &  & 87.61 & 9.86 & 98.87 & 47.89 & 100 & 99.44 & 100 & \textbf{100} & +1 \\
\hline
\textbf{Average} & 41.30 & 24.66 & 91.12 & 81.82 & 93.34 & 89.44 & \textbf{98.22} & \textbf{93.92} & -- &  & 44.99 & 16.28 & 75.05 & 46.90 & 98.35 & 95.10 & \textbf{98.95} & \textbf{97.23} & -- \\
\hline\hline
\end{tabular}
}
\end{table*}

\begin{table}
\renewcommand\arraystretch{1.2}
\centering
\caption{Ablation Study of SafeStep on Query Accuracy $\mathcal{A}_Q$ (GPT-oss:20B)}
\label{tab:ablation}
\setlength{\tabcolsep}{4pt}
\begin{tabular}{lccc}
\hline\hline
\textbf{Query task} & \textbf{OSP} & \textbf{w/o ToolBox} & \textbf{Safety4DM} \\
\hline
(1) ConditionType   & 33.1 & 34.9 & 100.0 \\
(2) RiskLevel      & 24.3 & 34.0 & 100.0 \\
(3) EI Count        & 90.5 & 90.8 & 98.3 \\
(4) EI Exist         & 58.0 & 56.3 & 100.0 \\
(5) Velocity        & 4.6  & 3.3  & 97.0 \\
(6) Acceleration     & 0.3  & 1.6  & 88.1 \\
(7) Heading          & 6.4  & 10.0 & 87.5 \\
(8) Color           & 25.7 & 23.0 & 91.2 \\
(9) Classification   & 21.2 & 18.5 & 90.9 \\
(10) Size            & 7.0  & 7.3  & 96.3 \\
(11) Status          & 0.0  & 0.0  & 87.0 \\
(12) Distance        & 1.4  & 1.4  & 94.5 \\
(13) OI Count        & 44.2 & 31.2 & 97.7 \\
(14) OI Exist        & 67.1 & 56.5 & 94.6 \\
(15) TTC             & 28.7 & 24.4 & 99.4 \\
(16) THW              & 25.1 & 29.3 & 99.1 \\
(17) SafeSpeed        & 9.0  & 4.5  & 97.9 \\
(18) StoppingDist.  & 28.5 & 24.5 & 98.6 \\
\textbf{Average}    & 34.8 & 33.7 & 96.3 \\
\hline\hline
\end{tabular}
\end{table}

\begin{table*}[]
\renewcommand\arraystretch{1.3}
\centering
\caption{Exploratory user survey of safety4dm (n = 35).}
\label{tab:user_survey}
\begin{tabular}{lllll}
\hline\hline
ID & Dimension             & Questionnaire item                                                                           & Mean±SD     & Agreement(\%) \\ \hline
Q1 & Information need      & I need more information about road risks than my current ADAS provides. & 4.31 ± 0.80 & 80.0          \\
Q2 & Perceived usefulness  & This function would help me understand risks on my current road.                             & 4.06 ± 0.84 & 80.0          \\
Q3 & Answer clarity        & The answers shown in the examples are easy to understand.                                    & 4.09 ± 0.95 & 77.1          \\
Q4 & Perceived reliability & I consider the answers shown in the examples to be reliable.                                 & 4.03 ± 0.71 & 82.9          \\
Q5 & Intention to use      & I would use this function if it were available in my vehicle.                                & 4.29 ± 0.79 & 80.0     \\    
\hline\hline
\end{tabular}
\end{table*}

\begin{figure}[h]
  \centering
  \includegraphics[width=0.4\textwidth]{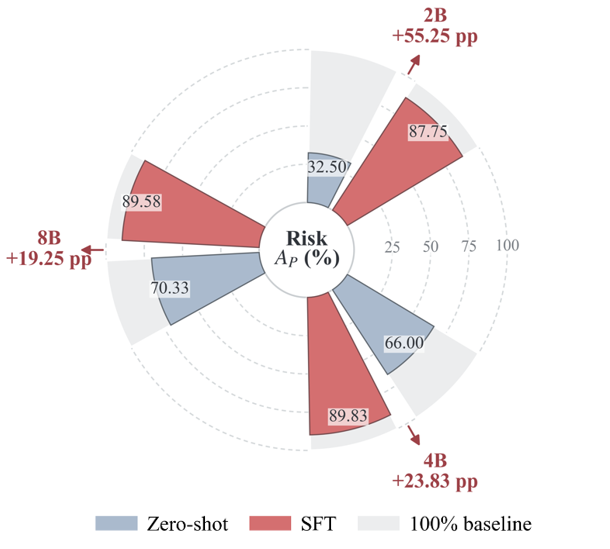}
  \caption{Overall perception accuracy before and after fine-tuning across three model scales.}
  \label{fig9}
\end{figure}

Figs.\ref{fig9} illustrates the effects of supervised fine-tuning (SFT) on road risk perception across different model scales and road conditions. As shown in Figs.\ref{fig9}, SFT substantially improves the joint risk perception accuracy of all three Qwen3-VL models, with gains of 55.25, 23.83, and 19.25 percentage points for the 2B, 4B, and 8B models, respectively. Notably, the fine-tuned 2B model outperforms the zero-shot 8B model, highlighting the effectiveness of domain-specific adaptation over simply increasing model size.

\begin{figure}[h]
  \centering
  \includegraphics[width=0.48\textwidth]{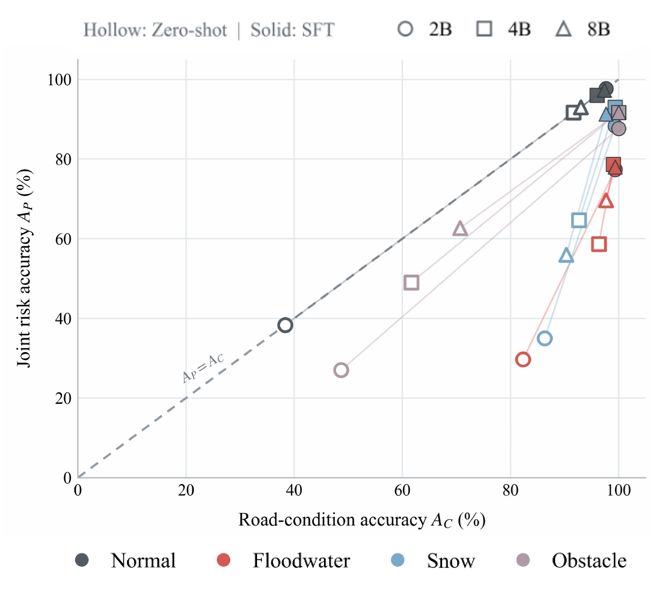}
  \caption{Road-condition classification accuracy versus joint risk perception accuracy across four road conditions.}
  \label{fig10}
\end{figure}

Figs.\ref{fig10} further reveals a persistent gap between road-condition classification accuracy and joint risk perception accuracy, particularly for floodwater scenarios. Although the fine-tuned models achieve classification accuracies exceeding 99\% for floodwater, their joint risk perception accuracies remain below 79\%. These findings suggest that identifying road hazards is considerably easier than accurately assessing their severity, indicating that fine-grained risk grading remains a key challenge for vision-language-based road safety perception.

\subsection{Generalization Across LLM Families}

To test how well SafeStep generalizes, we compare NLS query performance across different LLM families. For each family, we pick the largest model that runs on our hardware, and we use the Q4\_K\_M quantized version of every model. Table~\ref{tab:family} shows the overall results. SafeStep adapts smoothly to LLMs from different families. Except for Llama3.1:8B, all five models score above 98.10\% in $\mathcal{A}_C$ and above 93.40\% in $\mathcal{A}_Q$. We developed and refined the prompts with Qwen3:8B as the reference model, then applied them to the other families without any changes. These models still deliver competitive performance, which shows that SafeStep does not depend on the underlying model and generalizes well.

A closer look at the two metrics shows that the models diverge to different degrees in task understanding and query execution. For task understanding, all models except Llama3.1:8B (81.06\%) fall within a narrow $\mathcal{A}_C$ range of 98.10\%--98.95\%, less than one percentage point apart. This suggests that larger models have reached a ceiling in understanding the query tasks. For query execution, $\mathcal{A}_Q$ varies much more: DeepSeek-r1:32B performs best (97.23\%), and Llama3.1:8B performs worst (68.30\%). In terms of task bias, the Status task scores lowest ($\mathcal{B}_{\mathcal{T}_i} = -3$). This likely stems from semantic overlap at the task classification stage. For example, a model can read ``Is the vehicle ahead braking?'' as either a signal status query or an acceleration query, so its classification is less stable.

\begin{figure}
  \centering
  \includegraphics[width=0.5\textwidth]{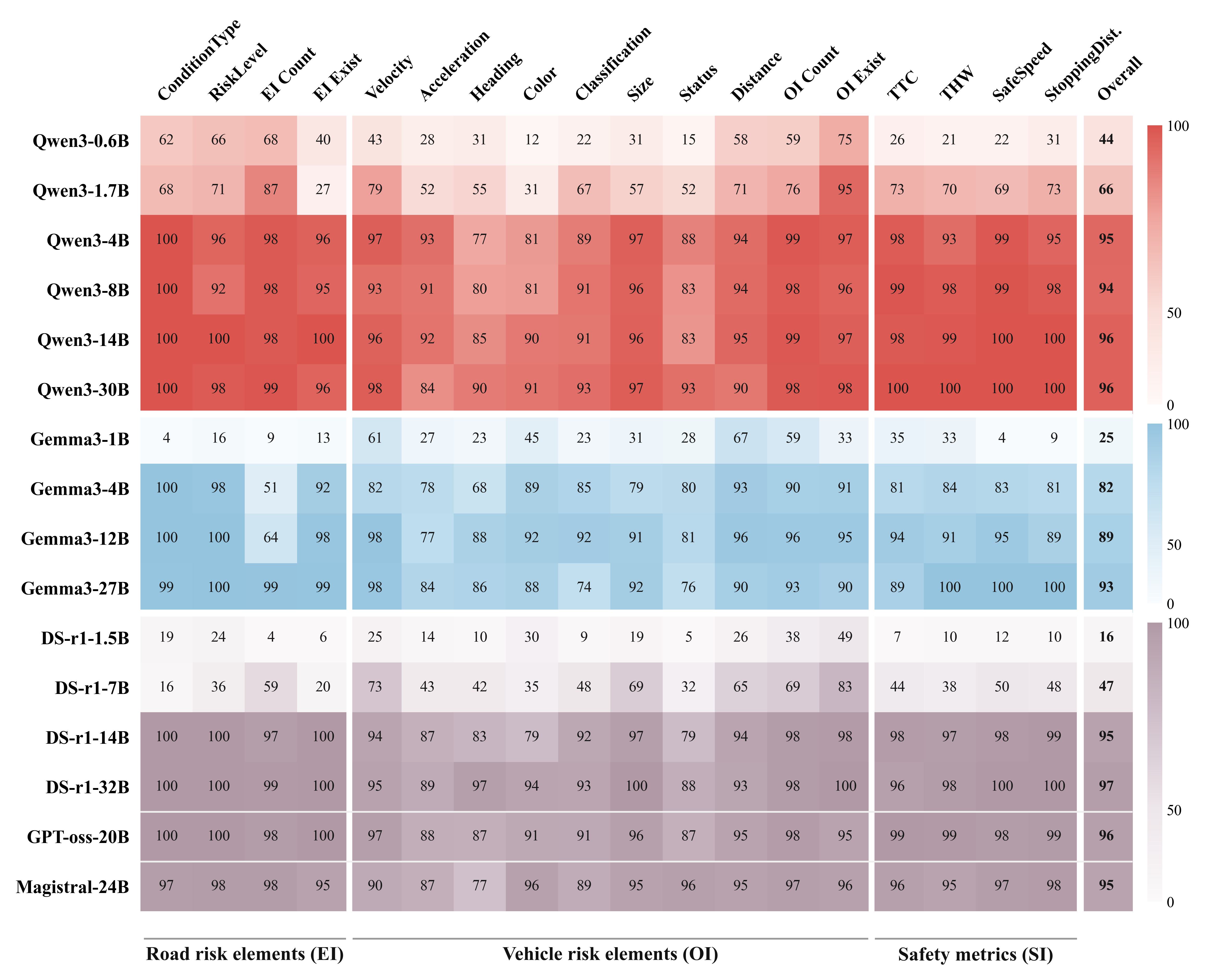}
  \caption{Heatmap of query accuracy $\mathcal{A}_Q$ (\%) of Safety4DM on each task across 17 LLMs. Colors distinguish model families, and darker shades indicate higher accuracy.}
  \label{fig_heatmap}
\end{figure}

\subsection{Effect of Model Scale}

In this experiment, we evaluate the query accuracy of Safety4DM across model sizes from 0.6B to 32B. We use three model families, Qwen3, DeepSeek-r1, and Gemma3, along with the GPT-oss and Magistral models. Table~\ref{tab:qwen} reports the results for the Qwen3 family, and Table~\ref{tab:gemma_deepseek} reports those for the Gemma3 and DeepSeek-r1 families. For clarity, Fig.~\ref{fig_heatmap} visualizes all the results. As Fig.~\ref{fig_heatmap}  shows, all three families follow a clear scaling trend: $\mathcal{A}_C$ and $\mathcal{A}_Q$ generally rise as model size grows. Notably, the three families cross the 90\% $\mathcal{A}_Q$ mark at very different sizes. Qwen3 already crosses it at 4B, DeepSeek-r1 needs 14B, and Gemma3 only gets there at 27B. This suggests that SafeStep needs a minimum model capacity to run reliably, and this minimum can differ several-fold across families.

Comparing across families, the performance gap is large at smaller sizes (below 14B). Qwen3:4B already reaches 94.66\% in $\mathcal{A}_Q$, while the larger DeepSeek-r1:7B reaches only 46.90\%. As model size grows, the gap closes quickly. At around 30B, the three families fall within an $\mathcal{A}_Q$ range of 93.92\%--97.23\%, less than four percentage points apart. Looking at each family in turn, Qwen3 hits its turning point earliest. From 1.7B to 4B, its $\mathcal{A}_Q$ jumps from 66.30\% to 94.66\%, a gain of 28.36 percentage points. After that, performance levels off: Qwen3:8B (94.45\%) and Qwen3:4B (94.66\%) perform about the same, and further scaling brings much smaller returns. The DeepSeek-r1 family is the most sensitive to model size. Its 1.5B model scores only 16.28\% in $\mathcal{A}_Q$, the lowest in all our experiments. The 7B model rises to 46.90\%, and the 32B model climbs to 97.23\%, the highest among all models we compare. The Gemma3 family shows the largest single-step gain of any family between 1B and 4B, with $\mathcal{A}_Q$ jumping from 24.66\% to 81.82\%. Growth then slows, and even its largest model, Gemma3:27B, reaches only 93.92\%.

\subsection{Ablation Study}

To test the role of each SafeStep component, we compare three configurations, all built on GPT-oss:20B. OSP uses a single prompt. It gives the model the role setting, the full language scene (LS), the question, and the output format constraints all at once, and the model generates the answer directly. w/o ToolBox keeps the two prompting stages of task classification and parameter extraction but does not call the query toolbox. Instead, it passes the resulting task type and query parameters to the model together with the LS, and the model filters the target entities and performs the numerical calculations on its own. Safety4DM uses the full SafeStep configuration.

As Table~\ref{tab:ablation} shows, OSP falls far behind the full SafeStep, with a gap of 61.5 percentage points in $\mathcal{A}_Q$. Even a large model with strong reasoning ability cannot give reliable answers to precise queries over structured scenes with a single prompt alone. SafeStep is therefore necessary for this kind of task. Moreover, w/o ToolBox reaches an average $\mathcal{A}_Q$ of 33.7\%, about the same as OSP (34.8\%). Without the query toolbox, the prompt decomposition in the first two SafeStep stages brings no real gain on its own. Task classification and parameter extraction exist to give the toolbox accurate inputs for its calls. Without the toolbox to carry out the query, the model still cannot retrieve and compute reliably in the scene, even when it correctly understands the query intent and extracts the right parameters.

At the task level, the two ablated configurations reach 67.1\% and 56.5\% in $\mathcal{A}_Q$ on the OI Exist task. The answer to this task is simply ``yes'' or ``no'', so a model can reach some accuracy by guessing, even without real spatial reasoning. In contrast, on tasks that require precise target localization and numerical calculation, both configurations drop sharply. They score 0.0\% on Status and below 5\% on Acceleration, Distance, and Velocity. This shows that LLMs struggle to locate targets and compute values directly in structured scenes, which further confirms the need for the query toolbox.

\begin{figure*}[h]
  \centering
  \includegraphics[width=0.9\textwidth]{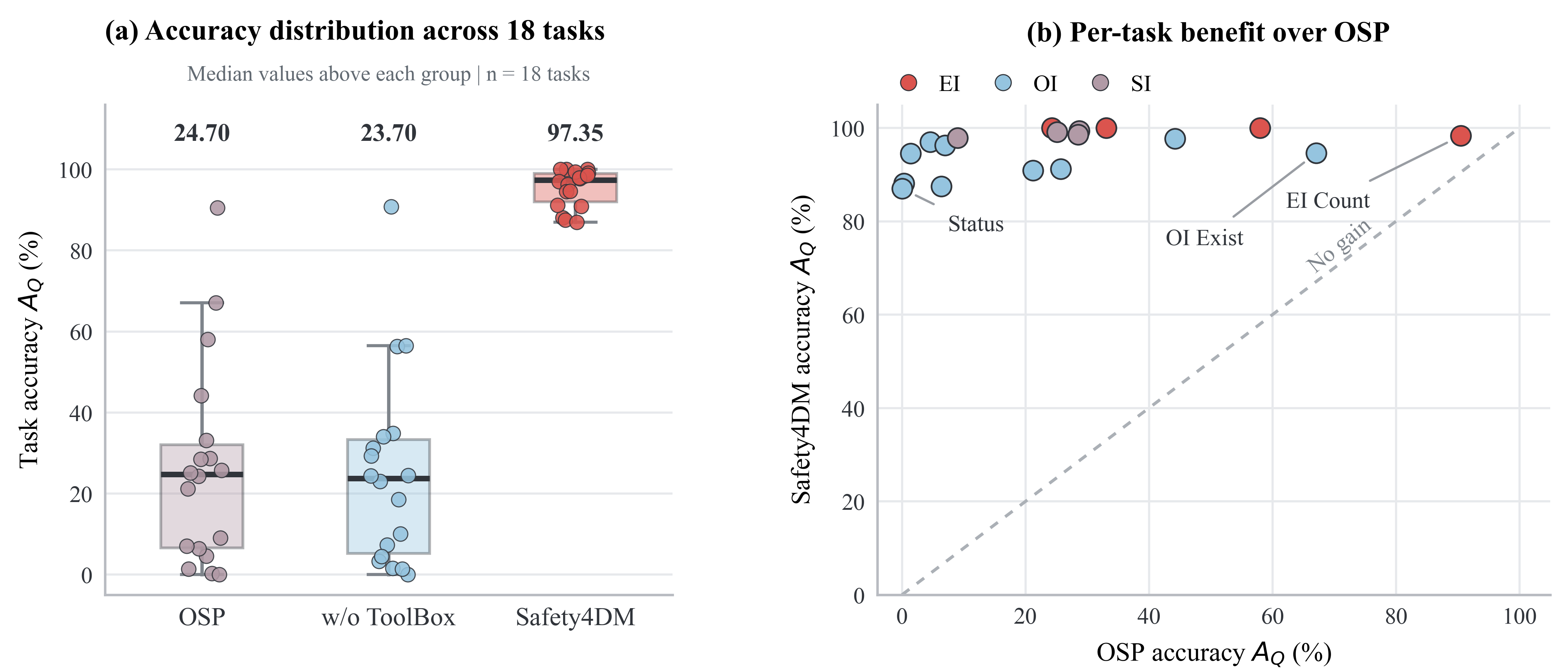}
  \caption{Ablation analysis of the SafeStep mechanism. (a) Distribution of query accuracy across 18 tasks under three configurations. (b) Per-task accuracy comparison between Safety4DM and one-step prompting (OSP).}
  \label{fig11}
\end{figure*}

Figs.\ref{fig11} presents the ablation results of the SafeStep mechanism across 18 safety-related query tasks. As shown in Figs.\ref{fig11}(a), the full Safety4DM configuration achieves consistently high task-level accuracy, with a median of 97.35\%, whereas one-step prompting (OSP) and the configuration without ToolBox exhibit substantially lower medians of 24.70\% and 23.70\%, respectively. Figs.\ref{fig11}(b) further demonstrates that Safety4DM outperforms OSP on all 18 tasks, with particularly pronounced improvements in vehicle attribute retrieval and safety metric computation. Notably, removing ToolBox reduces the sample-weighted overall query accuracy from 96.3\% to 33.7\%, which is comparable to the 34.8\% achieved by OSP. This observation indicates that step-wise prompt decomposition alone is insufficient for reliable querying of structured dynamic maps. Instead, the integration of LLM-based query interpretation with deterministic entity retrieval and numerical computation plays a critical role in achieving accurate and consistent safety-related responses.

\subsection{User Needs and Perceived Value}

We conducted an exploratory user survey with 35 respondents, all of whom had used driver assistance features in the past three months. Before seeing any system examples, respondents rated their need for road risk information (Q1). They then viewed the same set of 10 scenario-question-answer examples and rated Safety4DM on perceived usefulness, answer clarity, perceived reliability, and willingness to use (Q2–Q5). Every item used a five-point scale. Table~\ref{tab:user_survey} shows the results. Of all respondents, 80\% felt that current driver assistance systems provide too little road risk information, and 80\% felt the feature would help them understand risks on the road ahead. In addition, 77.1\% found the example answers easy to understand, and 80\% said they would use the feature if their vehicle offered it. Note that these results reflect only the perceived needs and acceptance of the respondents we recruited. They do not prove that Safety4DM actually improves driving safety.

% PRE-SURVEY PLANNING ONLY.
% All values below are synthetic, not observed survey results.
%
% Hypothetical response counts for ratings [1, 2, 3, 4, 5]:
% Q1: [1, 5, 18, 43, 33]
% Q2: [2,  5, 16, 47, 30]
% Q3: [1,  4, 11, 46, 38]
% Q4: [5, 12, 27, 40, 16]
% Q5: [4,  8, 17, 43, 28]
%
% Each row contains 100 hypothetical responses.
% Agreement = percentage of ratings 4 or 5.

\section{Conclusion}

This paper presents Safety4DM, a safety-oriented language system for VRC dynamic maps. We build RoadRisk, a real-world road image dataset, and fine-tune a large model on it for road safety risk perception. We also design SafeVRC-Sim, a VRC cooperative simulation framework, along with SafeDM-QA, a safety question-answering dataset. Finally, we propose SafeStep mechanism. Together, these components complete the pipeline from real-world visual risk perception to natural language safety queries. Experiments show that domain-specific fine-tuning brings much larger gains than scaling up model size: the fine-tuned Qwen3-VL-4B alone reaches 89.83\% accuracy in risk assessment. SafeStep transfers to LLMs from different families without any changes. With medium and large models (20B and above) from five families, it achieves over 93\% natural language query accuracy in every case. In future work, we will model how vehicles respond to road risks and further improve the accuracy of risk level grading.

%\section*{Acknowledgments}

\bibliography{bare_jrnl_new_sample4}
\bibliographystyle{IEEEtran}

\begin{IEEEbiographynophoto}{Zhengshu Zhou}
received the B.E. degree in software engineering from Tianjin University, Tianjin China, in 2013. He received the Master and Ph.D. degree in informatics from Nagoya University, Nagoya Japan, in 2020 and 2022, respectively. He is currently an Associate Professor with the College of Artificial Intelligence, Tianjin University of Science \& Technology (TUST). Previously, he was a system engineer with the Japanese automotive industry. And he was a researcher at Hitachi Yokohama Research Laboratory, Yokohama, Japan. His research interests mainly include software engineering and autonomous driving.
\end{IEEEbiographynophoto}

\begin{IEEEbiographynophoto}{Shaoyu Wang}
received the B.E. degree in computer science and technology
from North China Institute of Aerospace Engineering, Langfang, China, in
2024. He is currently pursuing the M.S. degree with Tianjin University of
Science and Technology, Tianjin, China. His current research interests include
deep learning and autonomous driving.
\end{IEEEbiographynophoto}

\begin{IEEEbiographynophoto}{Lu Tao}
is currently a tenure-track associate professor at Wuhan University.
He received his B.S. and M.E. degrees in Geographic Information Systems
and Navigation, Guidance, and Control from Wuhan University in 2014 and
2017, respectively. He obtained his Ph.D. degree in Informatics from Nagoya
University in 2023, and was then a researcher at the Institute of Innovation
for Future Society from 2023 to 2024. His research interests include vehicle
motion prediction, sensor fusion, high-definition maps, dynamic maps, edge
and cloud computing, and connected autonomous vehicles.
\end{IEEEbiographynophoto}

\begin{IEEEbiographynophoto}{Xinhui Hu}
received the B.E. degree in computer science and technology
from Hebei GEO University, Shijiazhuang, China, in
2025. She is currently pursuing the M.S. degree with Tianjin University of
Science and Technology, Tianjin, China. Her research interests include
deep learning and autonomous driving.
\end{IEEEbiographynophoto}

\begin{IEEEbiographynophoto}{Jiaqi Gu}
is currently pursuing the M.E. degree in software engineering at the College of Artificial Intelligence, Tianjin University of Science and Technology, Tianjin, China, where he received his B.E. degree in software engineering. His research interests include software engineering, deep learning, and autonomous driving.
\end{IEEEbiographynophoto}

\begin{IEEEbiographynophoto}{Tingting Zhao}
obtained her Ph.D. degree from Tokyo Institute of Technology, Japan. She is the Professor of Tianjin University of Science \& Technology, China. She is the visiting researcher at RIKEN Center for Advanced Intelligence Project (AIP), Tokyo, Japan. She has published more than 50 papers in leading international journals or conferences of machine learning domain, including Neural Computation, Neural Networks, NeurIPS, ECML/PKDD, IJCAI, ACML and PAKDD. She is in charge of several major projects as the project leader, such as of National Natural Science Foundation of China and the project of Scientific Research Foundation for the returned overseas Chinese scholars. Her research interest mainly include machine learning, robot control, and data mining.
\end{IEEEbiographynophoto}

\newpage

\vfill

\end{document}